\documentclass[10pt,twocolumn,letterpaper]{article}

\usepackage{wacv}              

\usepackage{graphicx}
\usepackage{amsmath}
\usepackage{amssymb}
\usepackage{booktabs}
\usepackage{algorithm}
\usepackage{algpseudocode}

\usepackage[dvipsnames,table]{xcolor}

\usepackage{pgf-pie}
\usepackage{tikz}
\usepackage{pgfplots}
\usepackage{pgfplotstable}
\pgfplotsset{compat=1.17}
\usetikzlibrary{pgfplots.fillbetween}

\usepackage{array}
\usepackage{multirow}
\usepackage{pifont}
\usepackage{subcaption}
\usepackage{makecell}
\usepackage{tabularx}

\newcommand{\cmark}{\ding{51}}%

\definecolor{darkgreen}{HTML}{006400}

\definecolor{score1}{HTML}{ffdae2}
\definecolor{score2}{HTML}{ffdad3}
\definecolor{score3}{HTML}{fbddc6}
\definecolor{score4}{HTML}{f0e1c2}
\definecolor{score5}{HTML}{e2e5c8}

\newcolumntype{Y}{>{\raggedright\arraybackslash}X}
\newcolumntype{C}[1]{>{\centering\arraybackslash}p{#1}}

\usepackage{amsmath}
\DeclareMathOperator*{\argmax}{argmax}

\definecolor{lightred}{rgb}{1,0.8,0.8}

\usepackage{mdframed}
\definecolor{bg}{rgb}{0.05,0.05,0.05}

\usepackage{soul}

\newcommand{\perfect}[1]{{\sethlcolor{score5}\hl{#1}}}
\newcommand{\good}[1]{{\sethlcolor{score4}\hl{#1}}}
\newcommand{\neutral}[1]{{\sethlcolor{score3}\hl{#1}}}
\newcommand{\bad}[1]{{\sethlcolor{score2}\hl{#1}}}
\newcommand{\worst}[1]{{\sethlcolor{score1}\hl{#1}}}

\definecolor{primary}{HTML}{6750A4}
\definecolor{secondary}{HTML}{958DA5}
\definecolor{tertiary}{HTML}{7894ae}
\definecolor{error}{HTML}{B3261E}

\definecolor{primarycontainer}{HTML}{EADDFF}
\definecolor{secondarycontainer}{HTML}{E8DEF8}
\definecolor{tertiarycontainer}{HTML}{cce5ff}
\definecolor{errorcontainer}{HTML}{F9DEDC}

\definecolor{onprimarycontainer}{HTML}{21005D}
\definecolor{onsecondarycontainer}{HTML}{1D192B}
\definecolor{ontertiarycontainer}{HTML}{001e31}
\definecolor{onerrorcontainer}{HTML}{001e31}

\usepackage{colortbl}
\newcolumntype{g}{>{\columncolor[gray]{0.8}}c}

\usepackage[pagebackref,breaklinks,colorlinks]{hyperref}

\usepackage[capitalize]{cleveref}
\crefname{section}{Sec.}{Secs.}
\Crefname{section}{Section}{Sections}
\Crefname{table}{Table}{Tables}
\crefname{table}{Tab.}{Tabs.}

\def\wacvPaperID{2075} 
\def\confName{WACV}
\def\confYear{2025}

\definecolor{red}{HTML}{FF0000}
\definecolor{green}{HTML}{00AA00}
\definecolor{purple}{HTML}{FF00FF}
\definecolor{blue}{HTML}{0000FF}

\begin{document}

\title{Human Pose Descriptions and Subject-Focused Attention for Improved Zero-Shot Transfer in Human-Centric Classification Tasks}

\author{Muhammad Saif Ullah Khan$^{1,2}$, Muhammad Ferjad Naeem$^{3}$, Federico Tombari$^{4}$, Luc Van Gool$^{3}$, \\
Didier Stricker$^{1,2}$, and Muhammad Zeshan Afzal$^{1,2}$ \\
\small{$^{1}$} \small{University of Kaiserslautern-Landau, Kaiserslautern, Germany} \\
\small{$^{2}$} \small{Deutsches Forschungszentrum für Künstliche Intelligenz, Kaiserslautern, Germany} \\
\small{$^{3}$} \small{ETH Zürich, Rämistrasse 101, Zurich, Switzerland} \\
\small{$^{4}$} \small{Technical University of Munich, Munich, Germany}
}
\maketitle

\begin{abstract}
  We present a novel LLM-based pipeline for creating contextual descriptions of human body poses in images using only auxiliary attributes. This approach facilitates the creation of the MPII Pose Descriptions dataset, which includes natural language annotations for 17,367 images containing people engaged in 410 distinct activities. We demonstrate the effectiveness of our pose descriptions in enabling zero-shot human-centric classification using CLIP. Moreover, we introduce the FocusCLIP framework, which incorporates Subject-Focused Attention (SFA) in CLIP for improved text-to-image alignment. Our models were pretrained on the MPII Pose Descriptions dataset and their zero-shot performance was evaluated on five unseen datasets covering three tasks. FocusCLIP outperformed the baseline CLIP model, achieving an average
  accuracy increase of 8.61\% (33.65\% compared to CLIP's 25.04\%). Notably, our approach yielded improvements of 3.98\% in activity recognition, 14.78\% in age classification, and 7.06\% in emotion recognition. These results highlight the potential of integrating detailed pose descriptions and subject-level guidance into general pretraining frameworks for enhanced performance in downstream tasks.
\end{abstract}

\section{Introduction}
\label{sec:intro}

Pretraining techniques that leverage multiple modalities have transformed deep learning, enabling models to capture intricate patterns from vast, unlabeled datasets~\cite{goyal2021self,gan2022vision,radford2021learning, naeem2023silc}. This approach is pivotal for zero-shot capabilities, which allow models to recognize concepts not seen during training~\cite{zhai2022lit,naeem2022i2dformer}. Within this evolving field, CLIP~\cite{radford2021learning} emerged as a significant advancement in language-image pretraining. It demonstrates impressive performance on various visual tasks by contrastively aligning images and text in a shared embedding space. CLIP-like methods~\cite{li2023scaling,jia2021scaling,yang2022attentive,zhai2022lit,dong2023maskclip} are effective but require a lot of data. The original CLIP model used 400 million image-text pairs because of its broad pretraining approach. Recent works~\cite{miao2022prior} have attempted to narrow the pretraining objective by using known priors. Following this, we propose using specialized text and a mechanism to limit the learning space when we are only interested in a related set of tasks, such as human-centric classification. This allows for enhanced zero-shot performance on unseen datasets using fewer pretraining samples compared to naive contrastive alignment.

\begin{figure}[t]
    \centering
    \includegraphics[width=\linewidth]{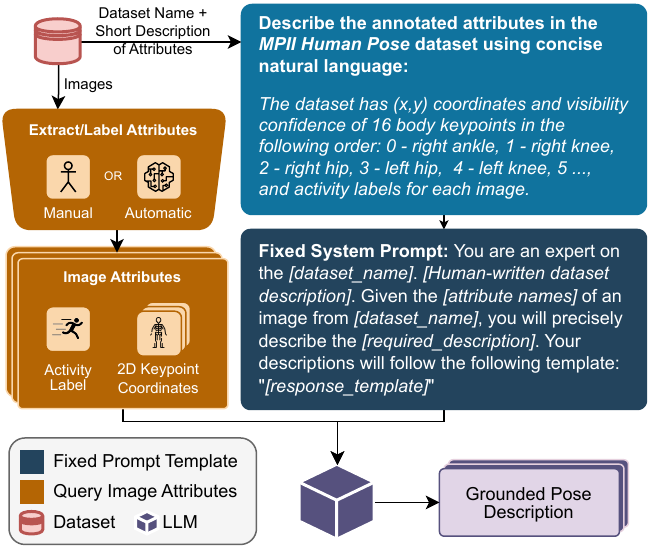}
    \caption{\textbf{Our LLM pipeline} creates grounded pose descriptions for images of people using only auxiliary attributes (activity labels and 2D keypoint coordinates) obtained from dataset annotations or extracted from the images using pretrained models.}
    \label{fig:prompting-method}
\end{figure}

We use the pipeline in~\cref{fig:prompting-method} to curate the MPII Human Pose Descriptions dataset. Our domain-specific text data contains around 14k samples with detailed natural language descriptions of activities and body postures.
Moreover, we draw inspiration from the human vision to further enhance the CLIP framework. The human gaze is characterized by rapid movements and fixations, where high-acuity vision is concentrated on limited spatial regions during fixation~\cite{foulsham2015eye,carrasco2011visual}. We create FocusCLIP by adding Subject-Focused Attention (SFA) to the vision side of the CLIP framework. This imitates the fixation stage of human vision, allowing the model to concentrate on task-relevant image regions. We demonstrate the effectiveness of our proposed pipeline comprising pose descriptions and SFA by performing a zero-shot evaluation of the pretrained FocusCLIP on several unseen human-centric tasks. Our main contributions include:

\begin{itemize}
    \item Introduction of a single-shot, structured LLM-prompting method for describing images in datasets by leveraging both class and image-level annotations.

    \item Public release of the MPII Pose Descriptions dataset comprising high-quality textual descriptions of human body posture and activities.

    \item Integrating subject-focused attention in generic contrastive pretraining via ROI heatmaps, establishing a novel paradigm for focused embedding learning.
    
    \item Superior zero-shot performance on three human-centric tasks compared to CLIP, including single-image activity recognition, age group classification, and emotion recognition. 
\end{itemize}

Our approach emphasizes task-related image features while retaining zero-shot capabilities, providing a promising direction for enhancing performance in applications requiring specialized knowledge with less pretraining data.

\begin{figure}[t]
    \centering
    \begin{tikzpicture}
        \begin{axis}[
            scale only axis,
            ybar,
            xtick=data,
            xticklabels={Activity, Age, Emotion, Mean},
            xticklabel style={font=\scriptsize},
            yticklabel style={font=\scriptsize},
            enlarge x limits=0.1,
            legend style={at={(0.5,-0.35)}, anchor=north,legend columns=-1, font=\scriptsize},
            bar width=5pt,
            ymin=0,
            grid=major,
            width=0.9\linewidth,
            height=1.5cm,
        ]
        \addplot[
            fill=gray,
            draw=gray
        ] coordinates {
            (0, 4.125) 
            (1,29.9975) 
            (2,30.6333) 
            (3,19.7614)
        };

        \addplot[
            fill=tertiary,
            draw=tertiary
        ] coordinates {
            (0, 6.49) 
            (1,37.16) 
            (2,31.48) 
            (3,25.04)
        };

        \addplot[
            fill=primary,
            draw=primary
        ] coordinates {
            (0,10.47)
            (1,51.94) 
            (2,38.54) 
            (3,33.65)
        };

        \legend{Chance, CLIP, FocusCLIP}
        \end{axis}
    \end{tikzpicture}
    \caption{FocusCLIP outperforms the baseline CLIP model on three zero-shot classification tasks (activity, age, emotion). Both models are pretrained on our MPII Pose Descriptions dataset.}
    \label{fig:focusclip_summary}
\end{figure}
\section{Related Work}
\label{sec:related_work}

\noindent \textbf{Vision-Language Models (VLMs)} \hspace{2pt} Multimodal learning~\cite{li2023scaling,jia2021scaling,yang2022attentive,zhai2022lit,dong2023maskclip,caron2021emerging,naeem2022i2dformer,yang2022vision,zhang2021vinvl,zhai2023sigmoid, naeem2023silc} has seen significant advancements, primarily focused on aligning and integrating textual and visual data. Despite the promising results of CLIP-like models~\cite{radford2021learning,li2023scaling,yang2022attentive,zhai2022lit,dong2023maskclip,xu2023demystifying,fang2023eva}, which map images and text into a shared embedding space, these models struggle with specialized tasks due to their reliance on generic pretraining data~\cite{miao2022prior}. Furthermore, their learning objective is global, with no mechanism for fine-grained alignment between image and text regions~\cite{naeem2022i2dformer,naeem2023i2mvformer}. Several works have attempted to address this limitation by introducing image-to-text cross-attention~\cite{naeem2022i2dformer}, random token masking~\cite{li2023scaling}, attentive token masking~\cite{yang2022attentive}, and masked self-distillation~\cite{dong2023maskclip}. Our work, FocusCLIP, explores another approach for improving fine-grained alignment between image and text by introducing an ROI heatmap during pretraining, explicitly guiding the focus toward task-relevant areas. This novel approach adds a degree of supervision to the self-supervised CLIP pretraining and enhances the alignment between text and visual data.

\noindent \textbf{Attention using Heatmaps} \hspace{2pt} Many VLMs develop an over-reliance on superficial language priors~\cite{selvaraju2019taking,peng2023empirical,salin2022vision}. Additionally, vision networks often focus on image areas that do not correlate with the regions humans look at when performing the same tasks~\cite{selvaraju2017grad, das2017human}. To address this, HINT~\cite{selvaraju2019taking} used human-generated attention heatmaps to guide model focus. These heatmaps served as explicit hints, showing which parts of an image humans considered essential for the task. The technique effectively improved the model performance on several tasks by providing a more reliable basis for grounding the predictions. Similarly,~\cite{rong2021human} also uses heatmaps of human gaze to highlight image regions humans deemed meaningful for bird classification. Notably, this use of heatmaps is distinct from the attention mechanisms commonly found in Transformer models~\cite{vaswani2017attention}. Using explicit human guidance to tune the model's focus has been similarly explored in various Visual Question Answering studies~\cite{lu2017knowing, ramakrishnan2018overcoming}. Our work introduces a conceptually similar attention map through the heatmap input. Integrating it into the CLIP framework sets our work apart, broadening applicability to multiple zero-shot tasks. Furthermore, unlike~\cite{selvaraju2019taking}, where heatmaps always need to be created manually, our proposed method does not require additional manual effort by human annotators in cases where keypoint annotations are already available.

\noindent \textbf{Large Language Models (LLMs) as Annotators} \hspace{2pt} In~\cite{su2022selective}, the authors explored the possibility of leveraging few-shot learning abilities of LLMs~\cite{brown2020language,anil2023palm,chowdhery2022palm,touvron2023llama,touvron2023llama2,openai2023gpt4,scao2022bloom} to generate new text datasets. Concurrently, other works demonstrated their capability to automate laborious annotation tasks~\cite{tekumalla2023leveraging, kuzman2023chatgpt}. Recently,~\cite{naeem2023i2mvformer} introduced a few-shot prompting strategy for writing natural language descriptions for images of animals and birds using LLMs. While~\cite{naeem2023i2mvformer} did not provide a framework for validating the text quality, they successfully used LLM-text to improve model performance. Drawing on this, we develop a prompting strategy in Section~\ref{sec:pose_captioning} for effectively describing images.

\begin{figure*}[ht]
    \centering
    \begin{subfigure}{0.50\linewidth}
        \begin{subfigure}{0.28\linewidth}
            \includegraphics[width=\linewidth]{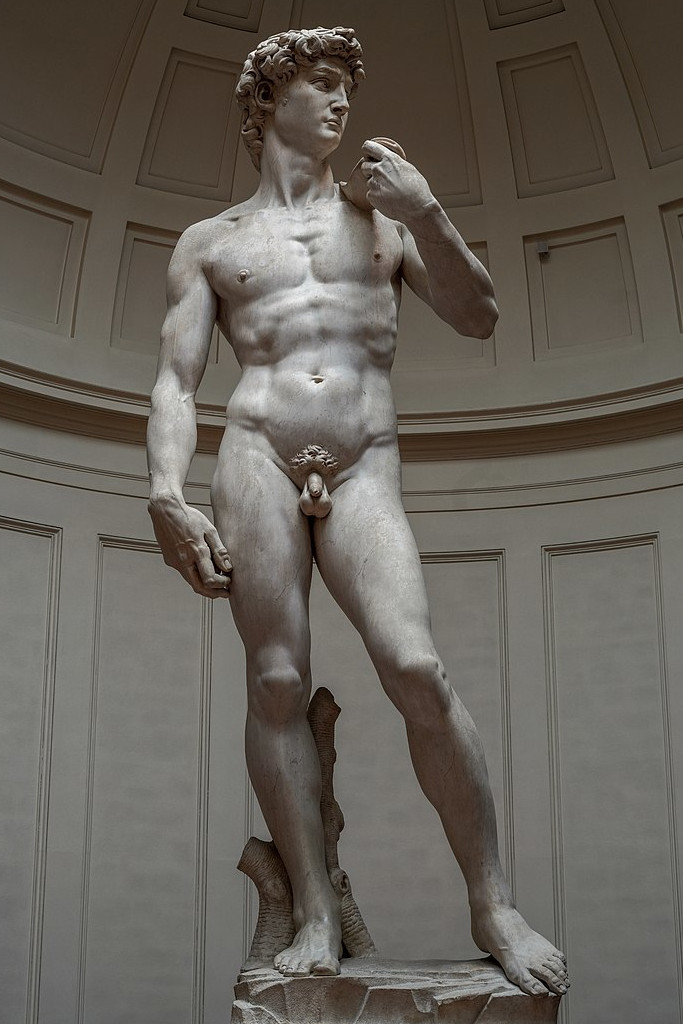}
            \begin{minipage}[c][0.5cm][c]{\linewidth}
                \centering
                \scriptsize
                \textbf{standing on a rock, holding a sling}
            \end{minipage}
        \end{subfigure}
        \hfill
        \begin{subfigure}{0.70\linewidth}
        \begin{mdframed}[innerleftmargin=4pt, innerrightmargin=4pt, innertopmargin=4pt, innerbottommargin=4pt]
            \scriptsize
            There is one person in the image who is standing on a rock, holding a sling. This activity is characterized by positioning on one's feet with one arm extended.
 
            The person in the center of the image is standing, which is evident from the location of the hips, knees, and ankles. The distribution of weight appears balanced, as both ankles are planted on the ground.
             
            The right leg of this person is slightly bent at the knee, with the ankle firmly on the ground, suggesting the leg might be bearing more weight. The left leg is also slightly bent at the knee, with the ankle grounded too, providing balance and support.
             
            The torso seems upright with a small inclination to the right, possibly for stability. The neck and head are erect, suggesting the individual could be focused on some target.
             
            The right arm is bent at an almost 90-degree angle at the elbow, with the wrist closer to the waist, implying the person is holding something, perhaps the sling's pouch. The left arm is markedly bent at the elbow, with a raised wrist level, suggesting the person might be reaching out or preparing to launch something with the sling.
             
            The head is looking straight forward or possibly slightly downward, indicating focus on the task at hand.
        \end{mdframed}
        \end{subfigure}
    \end{subfigure}
    \hfill
    \begin{subfigure}{0.48\linewidth}
    \centering
        \begin{subfigure}{0.28\linewidth}
            \includegraphics[width=\linewidth]{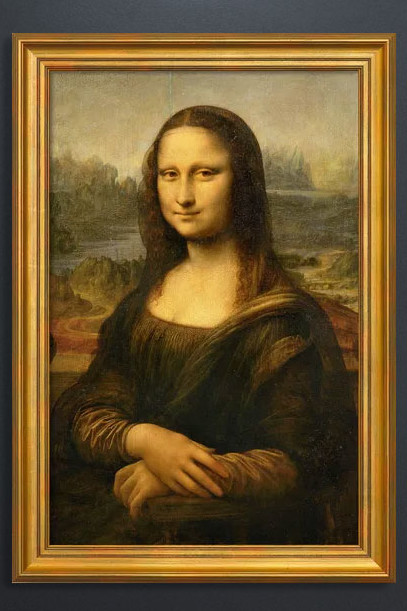}
            \begin{minipage}[c][0.5cm][c]{\linewidth}
                \centering
                \scriptsize
                \textbf{posing for a painting, smiling}
            \end{minipage}
        \end{subfigure}
        \hfill
        \begin{subfigure}{0.70\linewidth}
        \begin{mdframed}[innerleftmargin=4pt, innerrightmargin=4pt, innertopmargin=4pt, innerbottommargin=4pt]            \scriptsize
            There is one person in the image who is posing for a painting, smiling. The activity is characterized by a relaxed stance, with the person's body fairly upright and their expression visibly happy.
     
            The person is centrally positioned in the image. They are in a semi-seated pose, with their lower half obscured, implying they're sitting. The torso is slightly tilted to the right from center.
             
            Their right arm is bent at the elbow, extending down from the shoulder with the wrist located close to the hip near the pelvis. They may be supporting themselves in the seated position using their right hand.
             
            The left arm is outstretched away from the torso, forming a slight curve with a bent elbow. The wrist of their left hand is also at the same level as their right hand, which suggests it might be resting on a surface like an armrest.
             
            There's no visible information about the legs, as the keypoints for the knees and ankles return the "-1" value, hence they may be out of frame or obscured.
             
            The head is straight and upright with the neck-less curved. This, in conjunction with their activity, indicates that they might be looking directly towards the viewer, enhancing their smiling pose for the painting.
        \end{mdframed}
        \end{subfigure}
    \end{subfigure}
    \caption{\textbf{Sample Pose Description.} We use our pipeline to generate pose descriptions for two famous artworks, the Statue of David and the Mona Lisa. The LLM was provided body keypoints obtained using an off-the-shelf pose estimation network and manual activity labels.}
    \label{fig:samples_artworks}
\end{figure*}

\section{Methodology}
\label{sec:pose_captioning}

This section describes our novel pose descriptions dataset and the proposed FocusCLIP framework, which integrates Subject-Focused Attention (SFA) into CLIP~\cite{radford2021learning}.

\subsection{Pose Descriptions Dataset}

We introduce the \textbf{MPII Pose Descriptions Dataset} to improve zero-shot human classification, a collection of natural language annotations describing human body poses and activities. This dataset is derived from the MPII Human Pose dataset~\cite{andriluka14cvpr}, which provides detailed keypoint annotations for various human activities.

Using a novel LLM-driven pipeline (\cref{fig:prompting-method}), we automatically generated contextual descriptions of human poses based on auxiliary attributes such as 2D keypoint coordinates and activity labels. These attributes can either be obtained from the dataset annotations or extracted from images using pretrained models. By feeding this structured data into LLMs, we produced natural language descriptions that link body postures to human actions, enabling a more intuitive understanding of the data. We provide further details about prompt design in Appendix A.

The dataset contains 17,367 images, with 14,644 training and 2,723 validation samples. Each image includes up to four unique pose descriptions generated by state-of-the-art LLMs, providing diverse annotations for each sample. These descriptions (\cref{fig:samples_artworks} and \ref{tab:data-samples}) are grounded in image content and capture subtle posture and context variations beyond traditional keypoint annotations.

\begin{table*}[ht]
    \centering
    \scriptsize
    \begin{tabularx}{\linewidth}{cYcYcY}
        \toprule
        \raisebox{-0.8cm}{\includegraphics[width=0.1\linewidth]{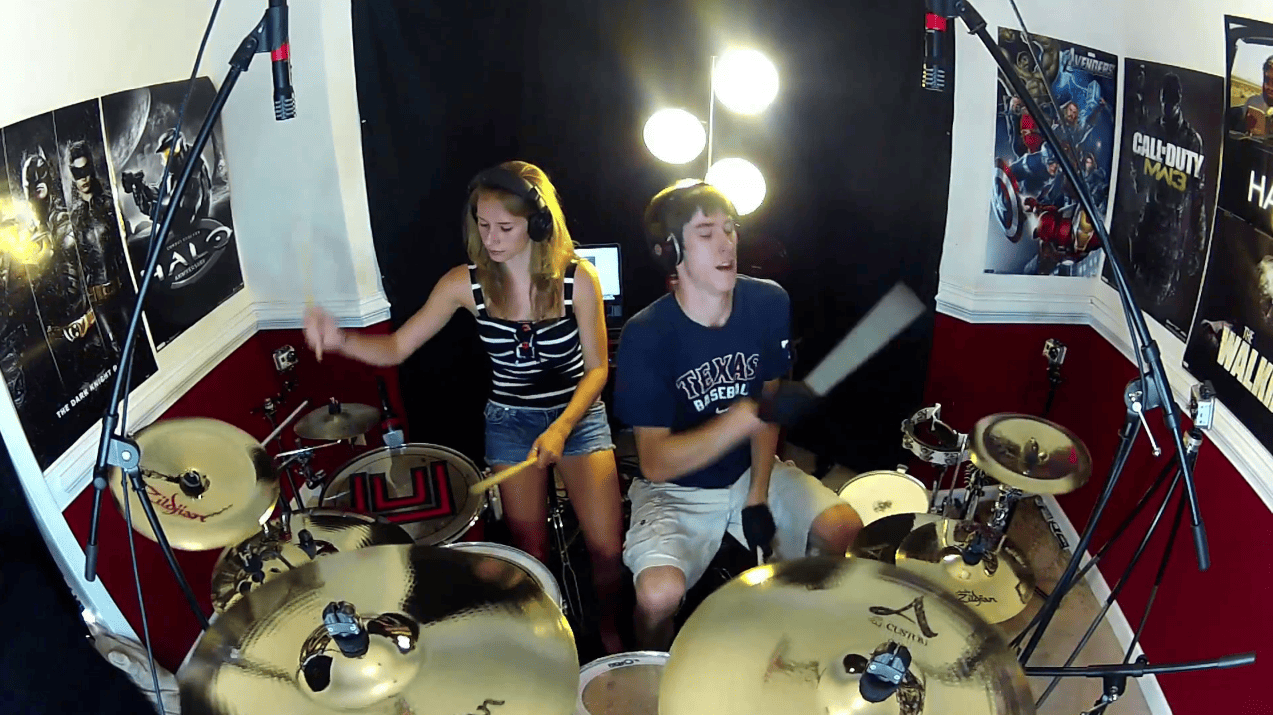}} &
        \cellcolor{score5} \textbf{GPT-3.5}: There are two people in the image who are playing music, specifically drums, while sitting. &
        \raisebox{-0.8cm}{\includegraphics[width=0.1\linewidth]{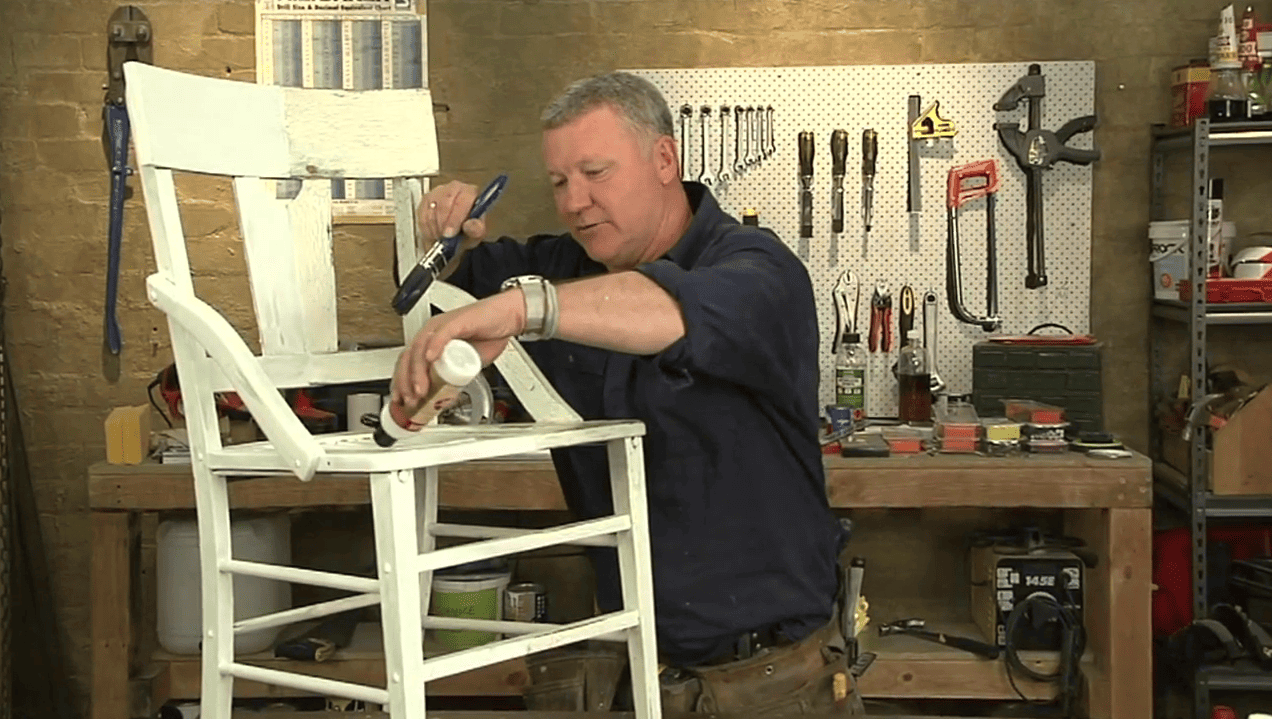}} &
        \cellcolor{score5} \textbf{GPT-3.5}: They have their right arm bent at the elbow, holding a paintbrush near their head. & 
        \raisebox{-0.8cm}{\includegraphics[width=0.1\linewidth]{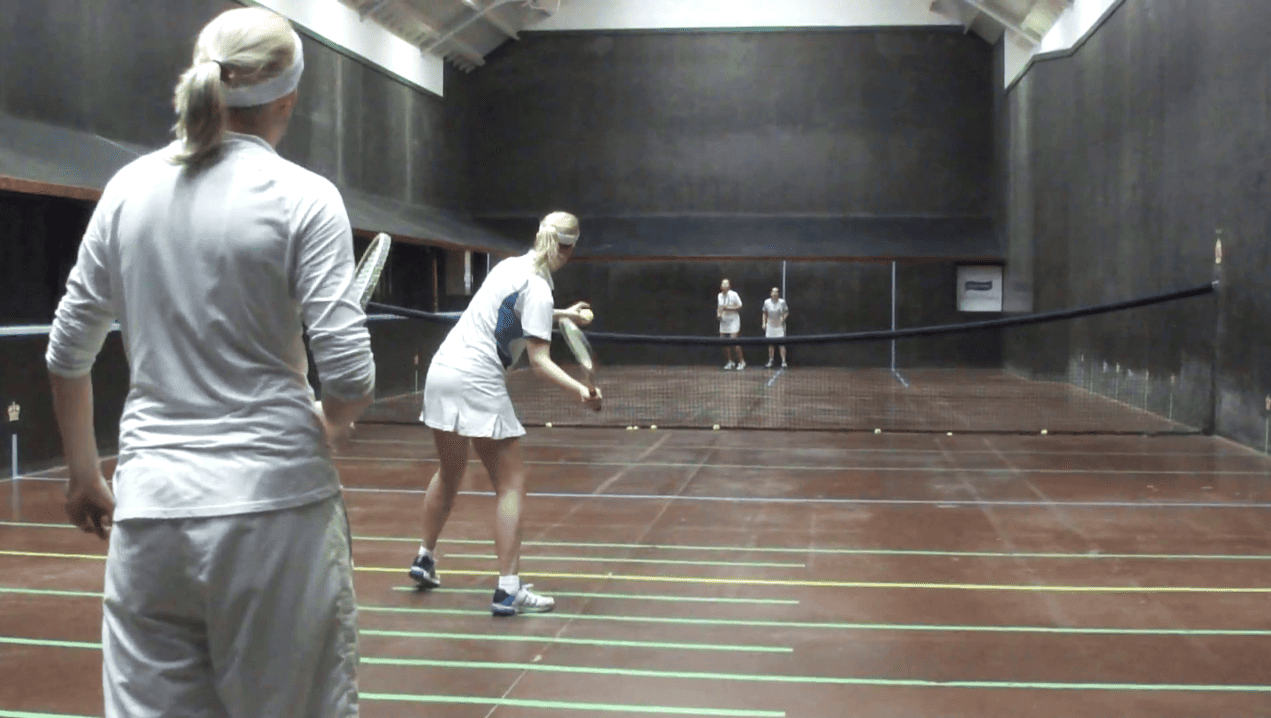}} &
        \cellcolor{score3} \textbf{GPT-3.5}: There are 2 people in the image who are playing doubles tennis. \\
        \midrule

        \raisebox{-0.8cm}{\includegraphics[width=0.1\linewidth]{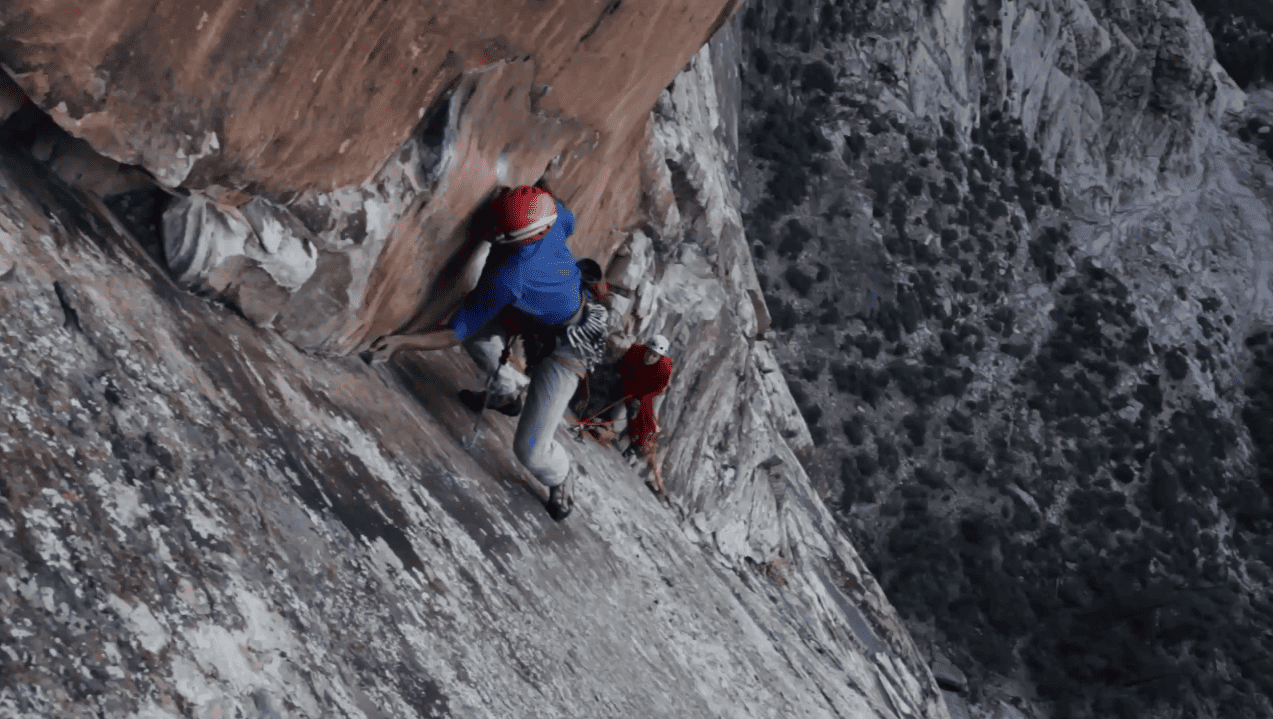}} &
        \cellcolor{score5} \textbf{GPT-3.5}: The torso appears to be slightly twisted to the right side while maintaining balance during climbing. &
        \raisebox{-0.8cm}{\includegraphics[width=0.1\linewidth]{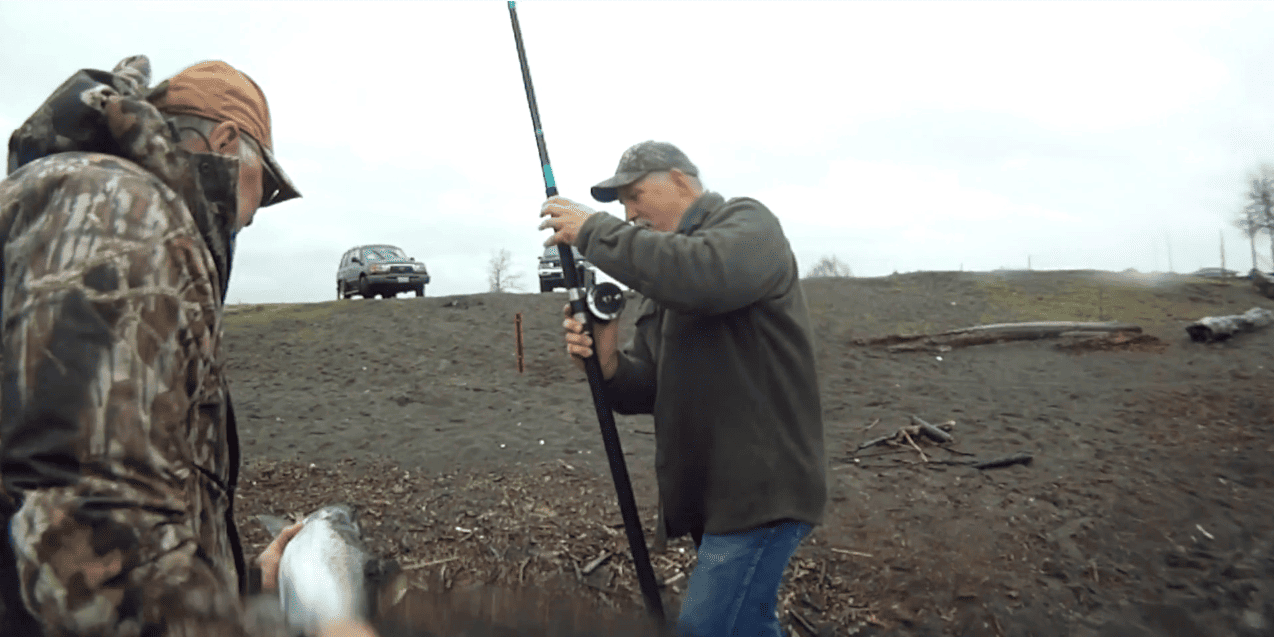}} &
        \cellcolor{score5} \textbf{GPT-3.5}: The right person is also standing with their left arm slightly bent and holding onto a fishing rod with their left hand. &
        \raisebox{-0.8cm}{\includegraphics[width=0.1\linewidth]{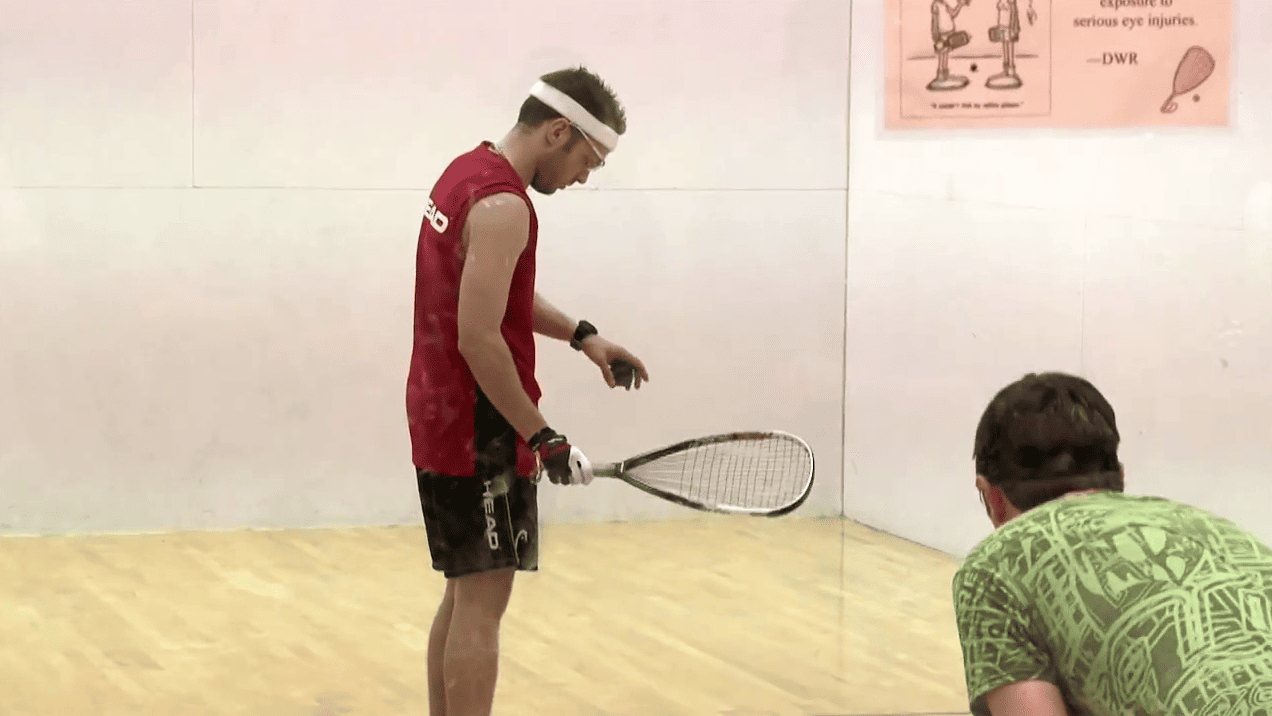}} &
        \cellcolor{score5} \textbf{GPT-3.5}: These limb positions suggest that this person might be preparing or executing a shot during their racquetball game. \\
        \midrule

        \raisebox{-0.8cm}{\includegraphics[width=0.1\linewidth]{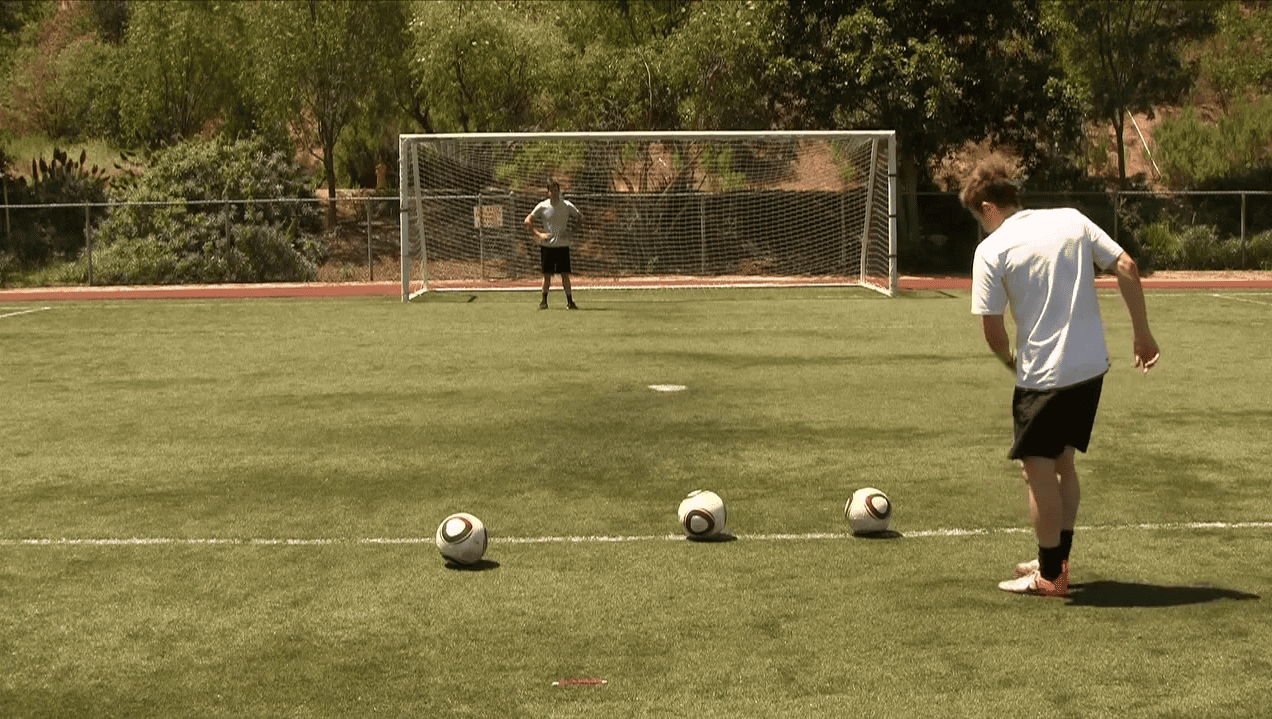}} &
        \cellcolor{score5} \textbf{GPT-4}: The left leg seems to be bearing most of their weight, as it's straightened and firmly planted on the ground. &
        \raisebox{-0.8cm}{\includegraphics[width=0.1\linewidth]{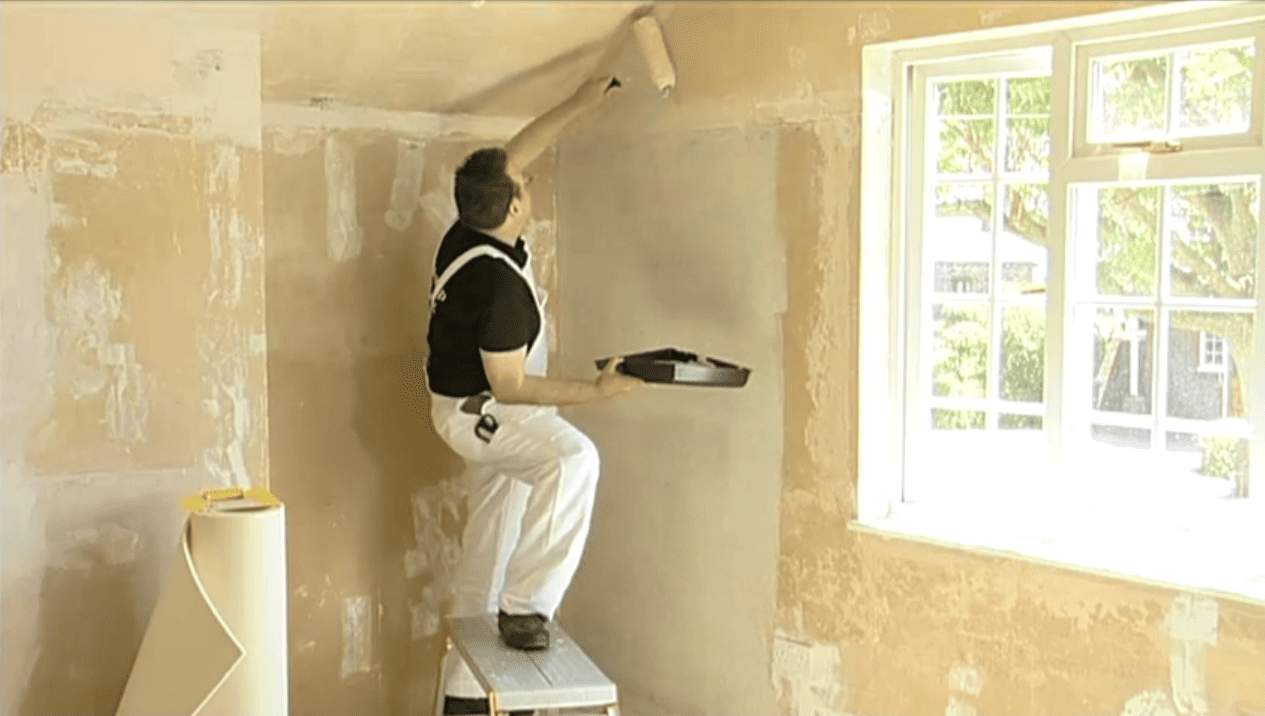}} &
        \cellcolor{score5} \textbf{LLaMa-2}: Their left arm is extended overhead, with their hand holding a tool or brush near the top of the wall. &
        \raisebox{-0.8cm}{\includegraphics[width=0.1\linewidth]{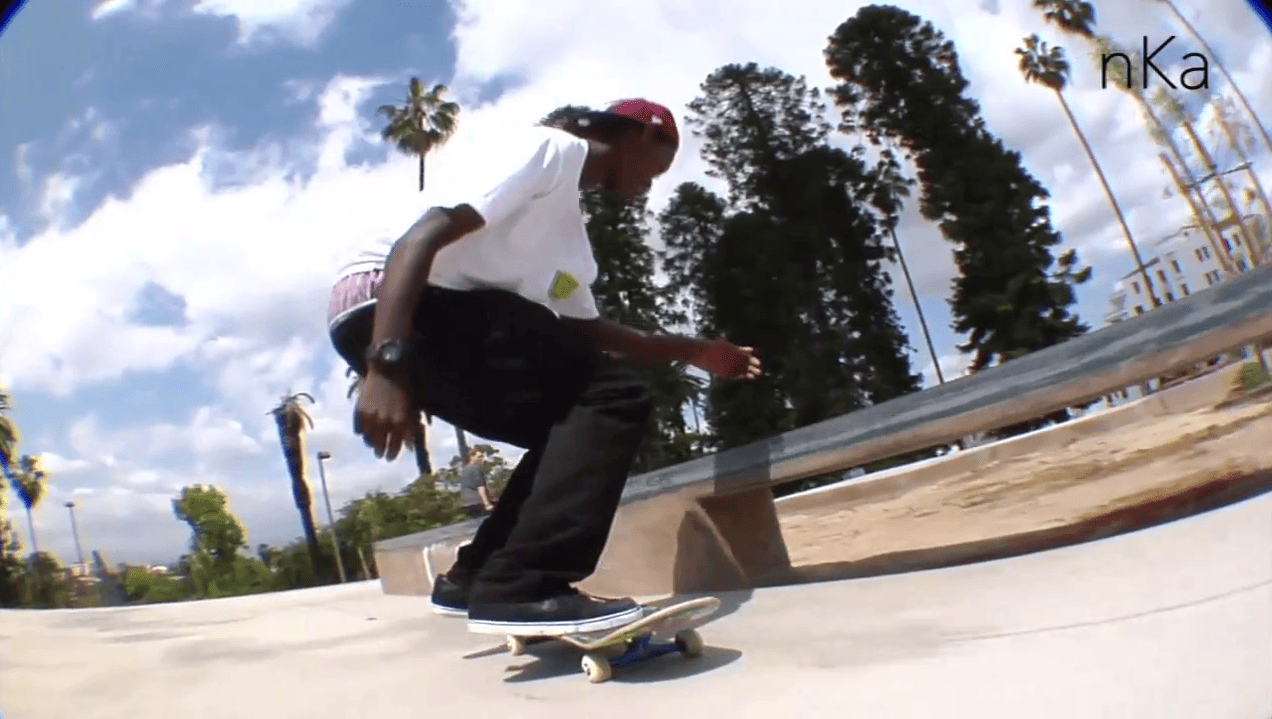}} &
        \cellcolor{score1} \textbf{LLaMa-2}: The person's left leg is extended behind them, with their foot pointing towards the ground. \\
        \midrule

        \raisebox{-1.1cm}{\includegraphics[width=0.1\linewidth]{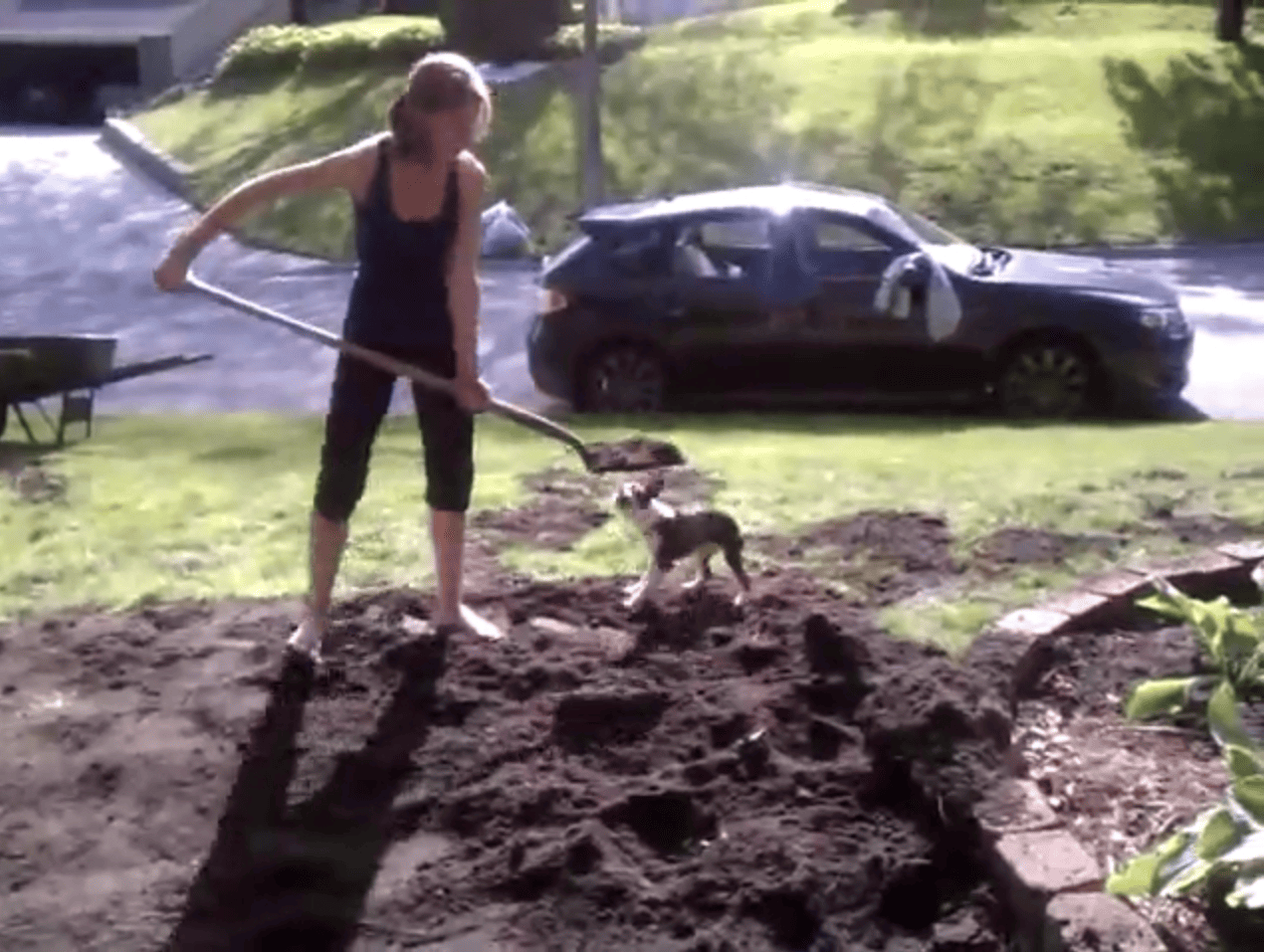}} &
        \cellcolor{score5} \textbf{GPT-4}: The person's right arm seems to be extended outwards and slightly downwards, possibly holding onto a tool such as a shovel handle. &
        \raisebox{-0.8cm}{\includegraphics[width=0.1\linewidth]{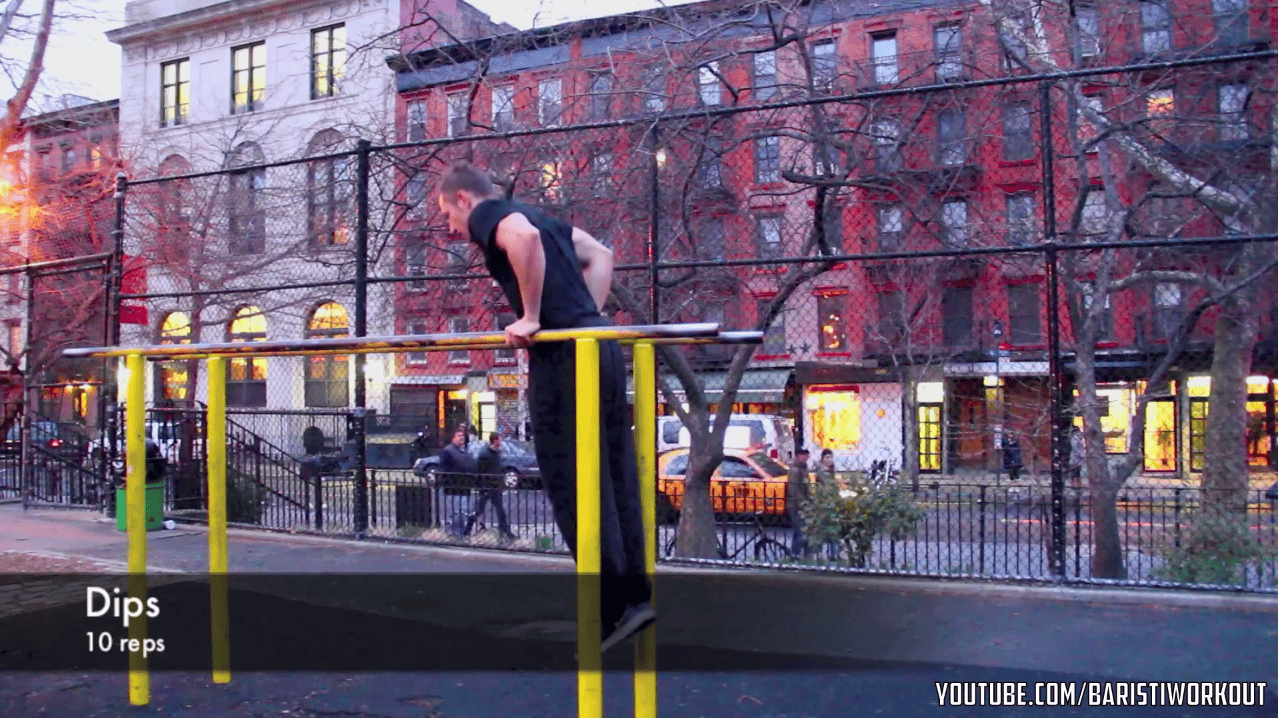}} &
        \cellcolor{score5} \textbf{GPT-4}: Lastly, their head seems to be tilted downwards possibly focusing on maintaining balance during this complex exercise routine. &
        \raisebox{-0.8cm}{\includegraphics[width=0.1\linewidth]{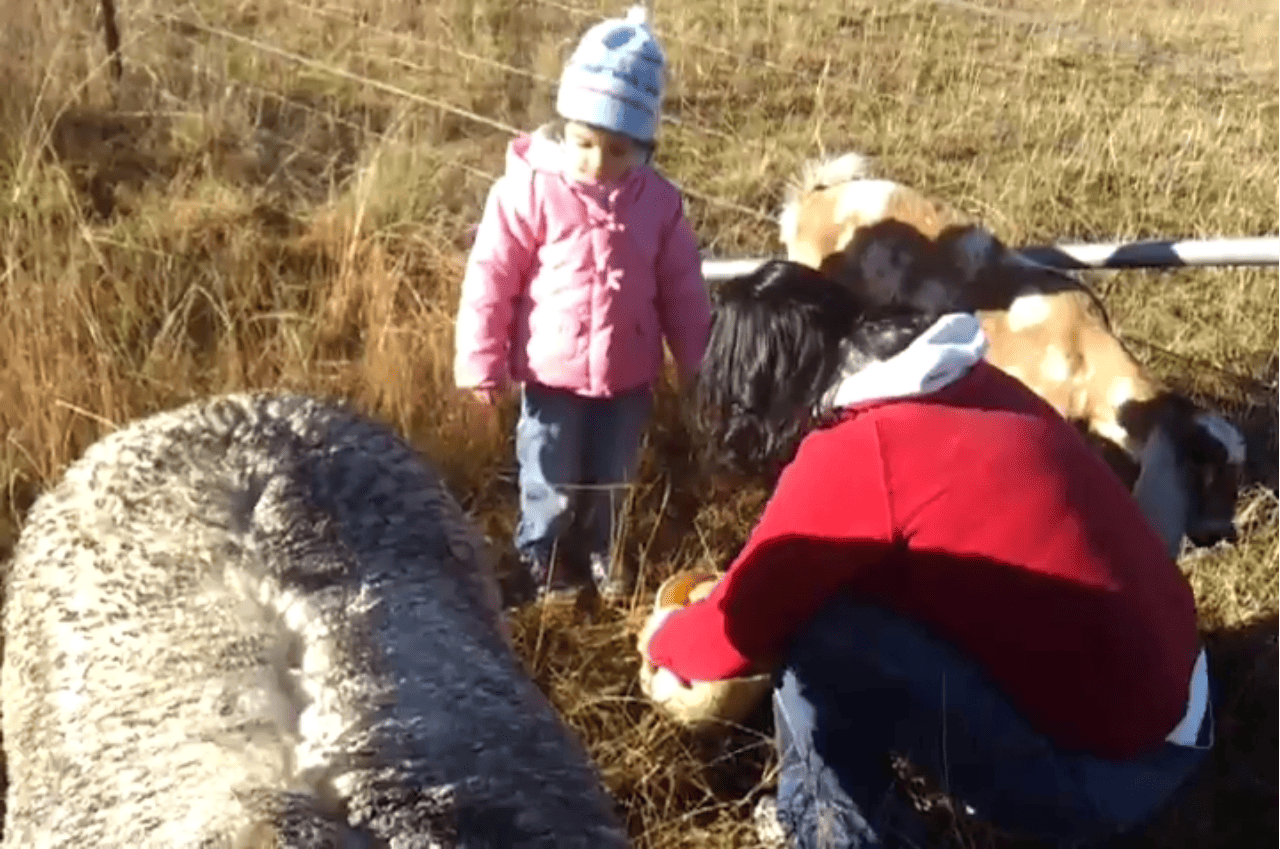}} &
        \cellcolor{score5} \textbf{GPT-3.5}: The second person is positioned towards the right side of the image, with a larger scale. \\
        \midrule

        \raisebox{-0.8cm}{\includegraphics[width=0.1\linewidth]{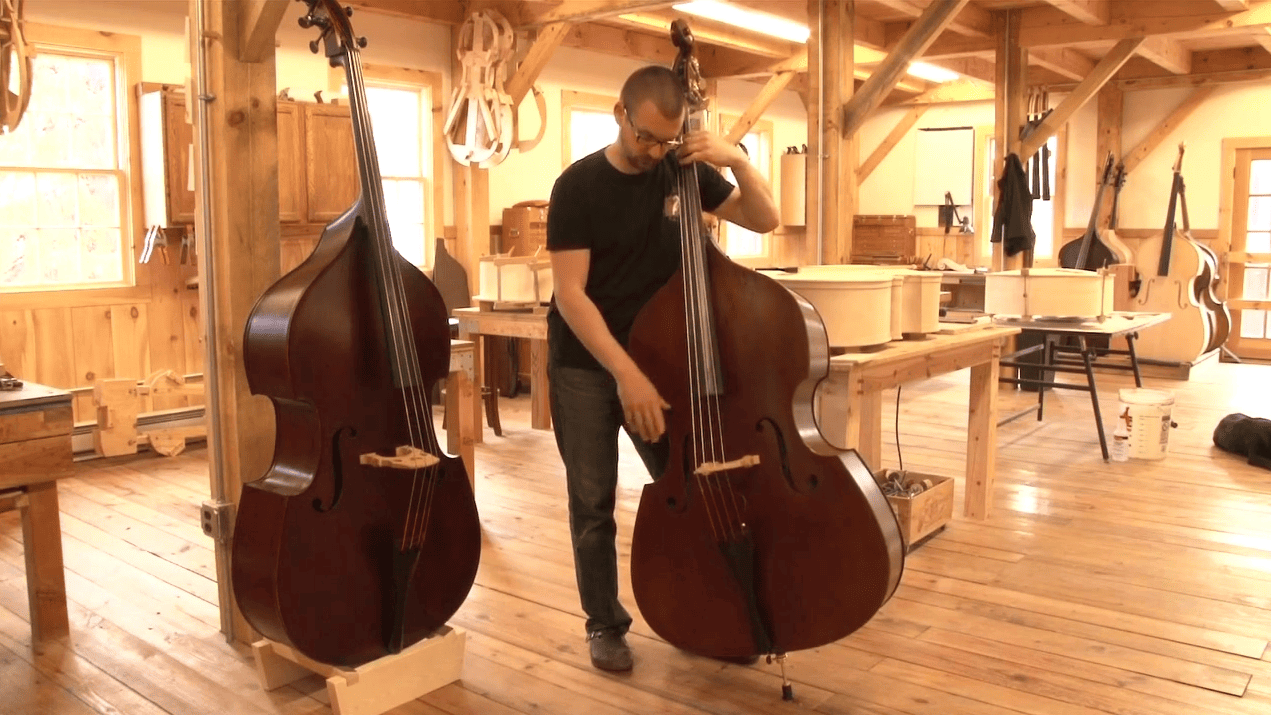}} &
        \cellcolor{score5} \textbf{GPT-4}: Their right arm is extended downwards with elbow bent at an angle greater than 90 degrees, suggesting that it's possibly resting or pressing against something - likely the double bass. &
        \raisebox{-0.8cm}{\includegraphics[width=0.1\linewidth]{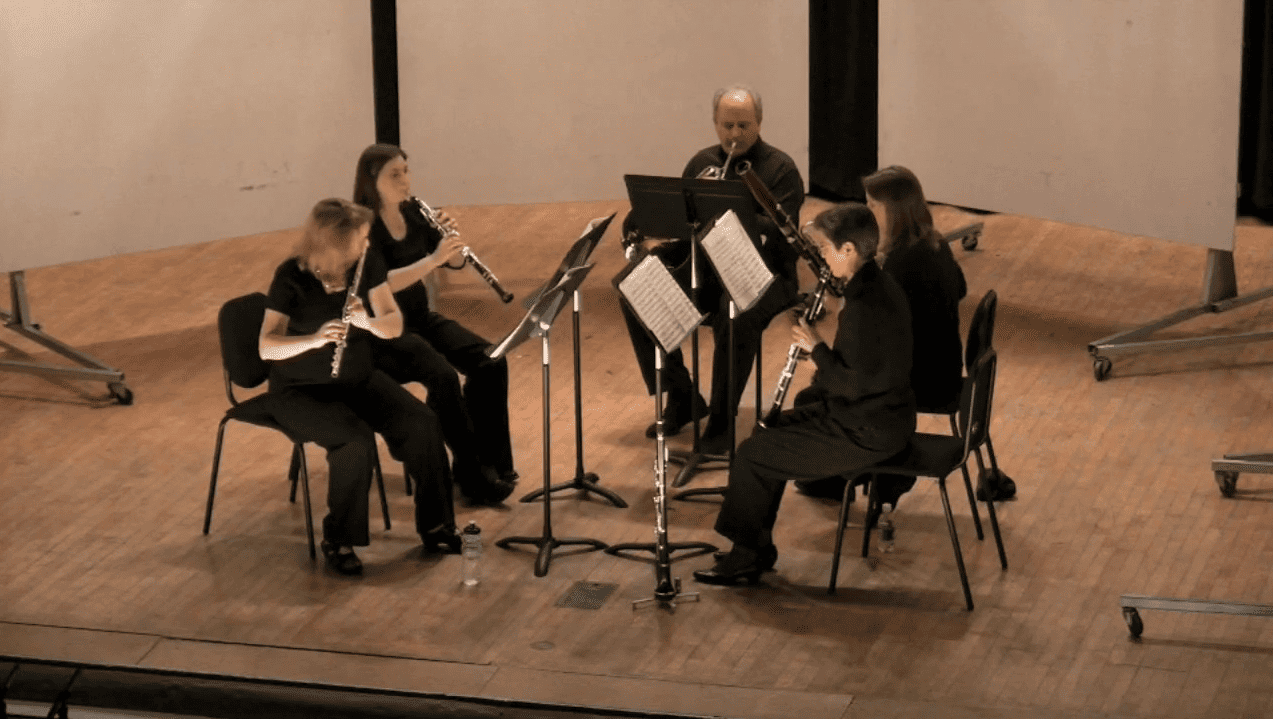}} &
        \cellcolor{score4} \textbf{GPT-3.5}: Overall, these individuals exhibit focused posture while playing music, leaning slightly forward to engage with their instruments and maintain balance during sitting position &
        \raisebox{-0.8cm}{\includegraphics[width=0.1\linewidth]{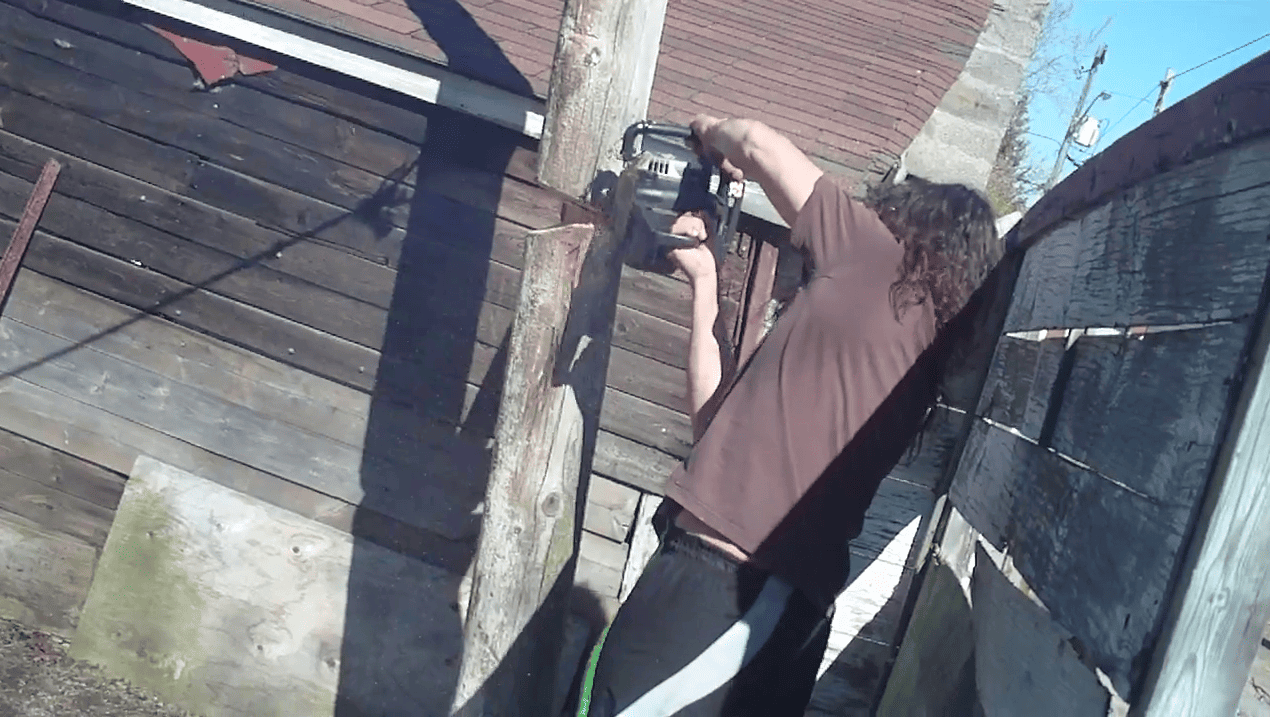}} &
        \cellcolor{score4} \textbf{LLaMa-2}: The person's pose suggests that they are in the process of trimming a branch on the right side of the shrub or tree, and their body is positioned at an angle to allow them to reach the desired area. \\
       
        \bottomrule
    \end{tabularx}
    \captionof{figure}{\textbf{Sample sentences from generated pose descriptions} reflect the LLM's nuanced understanding of activities and contextual relevance, offering precise interpretations of body language without seeing the image.}
    \label{tab:data-samples}
\end{table*}

\begin{figure*}[ht]
    \centering
    \begin{subfigure}[t]{0.29\linewidth}
        \includegraphics[width=0.93\linewidth]{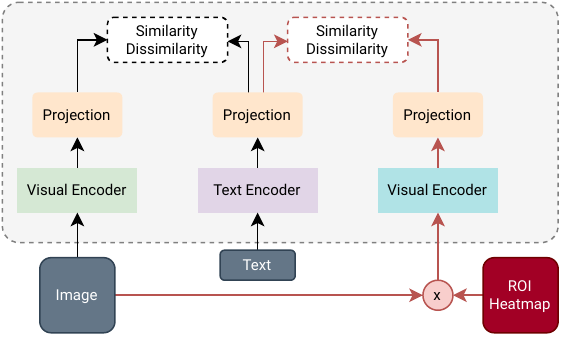}
        \caption{SFA in CLIP}
        \label{fig:focusclip}
    \end{subfigure}
    \hfill
    \begin{subfigure}[t]{0.29\linewidth}
        \includegraphics[width=\linewidth]{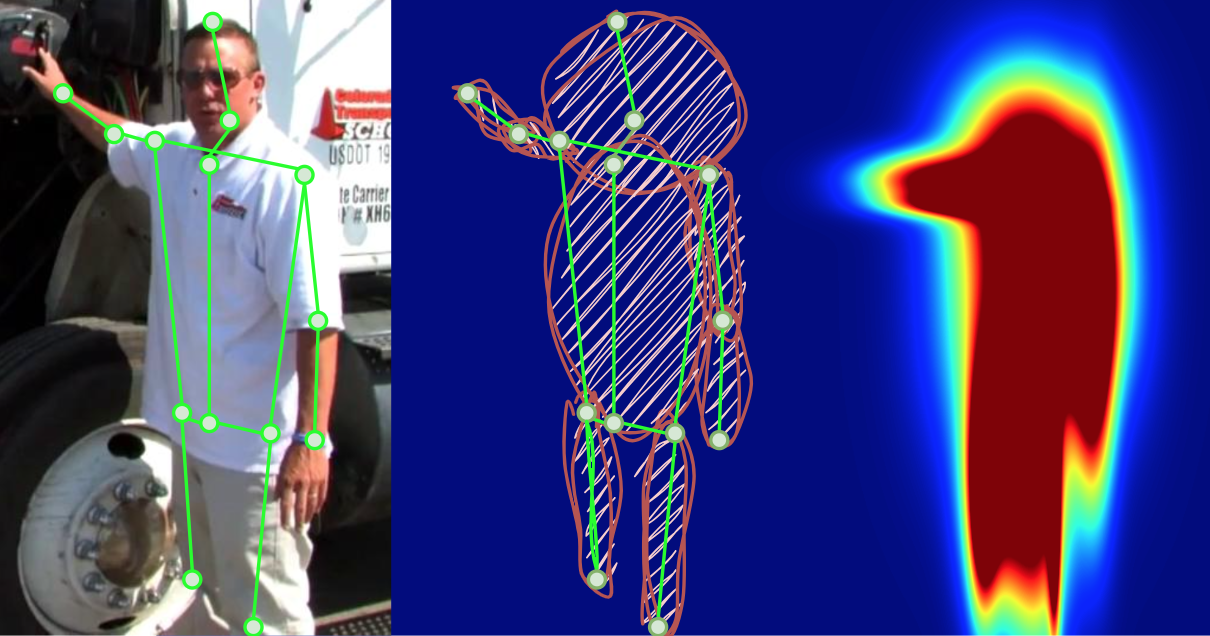}
        \caption{Heatmap Generation Process}
        \label{fig:heatmap_creation}
    \end{subfigure}
    \hfill
    \begin{subfigure}[t]{0.39\linewidth}
        \includegraphics[width=\linewidth]{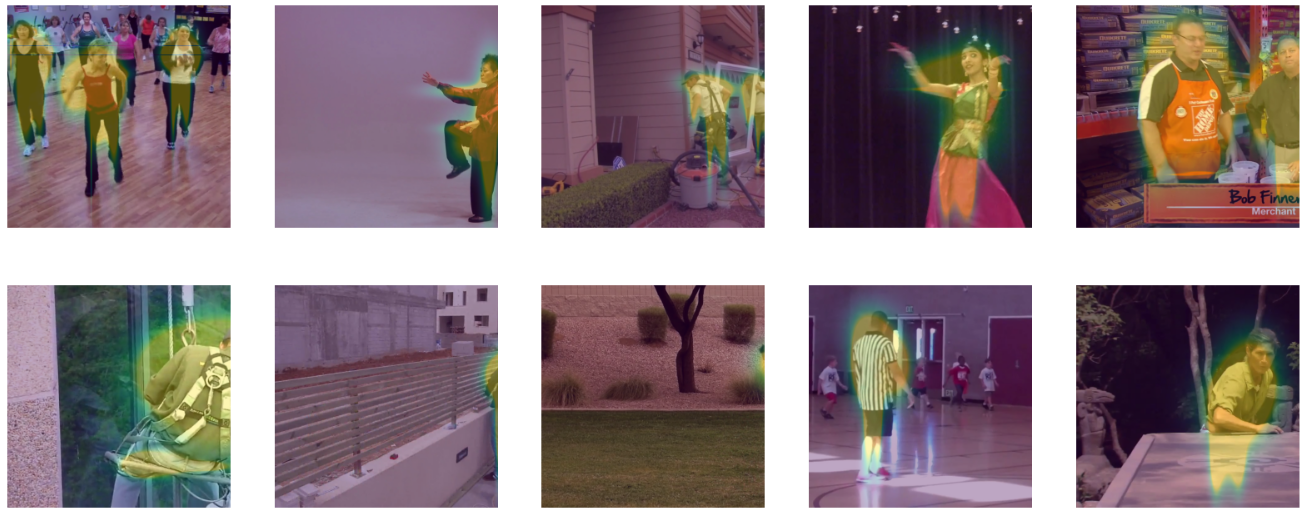}
        \caption{Heatmap Examples}
        \label{fig:heatmap_examples}
    \end{subfigure}
    \caption{\textbf{The FocusCLIP framework.} (a) We add SFA in CLIP, highlighting important areas for downstream tasks, and train the model with a dual contrastive loss to learn a joint embedding space between the raw image, heatmap-highlighted image, and text. The text consists of pose descriptions for people in images written by an LLM. (b) The heatmap generation process uses keypoints, which involves grouping keypoints into body parts and blending them together using Gaussian ellipses to produce the heatmaps. (c) This technique effectively highlights individuals within diverse environments.}
\end{figure*}

\subsubsection{Applications and Impact}

This dataset designed to bridge the gap between visual pose estimation and natural language understanding, making it valuable for various human-centric applications:

\noindent \textbf{Multimodal Learning}: The dataset facilitates joint learning between visual and textual modalities, enabling models to better align human pose with contextual descriptions.

\noindent \textbf{Activity Recognition}: By incorporating rich textual descriptions of human activities, the dataset enhances models' ability to recognize and classify actions from image and text inputs, improving zero-shot generalization.

\noindent \textbf{Fine-Grained Pose Analysis}: The detailed descriptions help models distinguish subtle differences in body postures, supporting tasks like behavior analysis, sports activity classification, and human-object interaction recognition.

\noindent \textbf{Text-Enhanced Keypoint Estimation}: These contextual descriptions offer additional information for improving the precision and interpretability of pose estimation models, especially in ambiguous scenarios.

Beyond pose estimation, the dataset can also be used for zero-shot human classification tasks by providing a deeper understanding of human behavior from visual and linguistic perspectives. We demonstrate this with our FocusCLIP framework.
\subsection{FocusCLIP Framework}
\label{sec:method}

Let \( \mathcal{D} = \{(I_i, T_i, H_i) \mid 1 \leq i \leq N \} \) be our training dataset, containing $N$ samples. In this dataset, \( I_i \) represents the input image, \( T_i \) denotes the corresponding textual description of the human pose, and \( H_i \) is the associated ROI heatmap for each $i$-th sample. Let $\phi_I: I \mapsto E_I$, $\phi_H: H \mapsto E_H$, and $\phi_T: T \mapsto E_T$ be encoding functions that map the image, heatmap, and text, respectively, into an embedding space. The objective is to learn a function \( f: (E_I, E_H) \mapsto E_T \) that maps the visual embeddings \( E_I \) and \( E_H \) to the textual embedding \( E_T \). We minimize a dual contrastive loss function $\mathcal{L}_{\text{SFA}}$ that quantifies the distance between matched and unmatched \( (E_I, E_H, E_T) \) triplets.

\subsubsection{Subject-Focused Attention (SFA)}
\label{sec:encoders}

The SFA mechanism enhances FocusCLIP by incorporating an additional encoder (\cref{fig:focusclip}). This encoder leverages Region of Interest (ROI) heatmaps to highlight salient regions—such as human figures—within the input images. This mechanism aims to improve model attention to task-relevant areas while maintaining alignment with corresponding textual descriptions.

Given an input image \( I \), its ROI heatmap \( H \) is generated based on keypoint annotations (\cref{fig:heatmap_creation}). The heatmap is applied through element-wise multiplication to form a masked image \( I_H = I \odot H \), selectively emphasizing important regions. The visual encoder \( \phi_I \), the text encoder \( \phi_T \), and the ROI encoder \( \phi_R \) transform the image \( I \), the text \( T \), and the masked image \( I_H \), respectively, into their corresponding embeddings \( E_I \), \( E_T \), and \( E_H \).

The SFA mechanism optimizes a dual contrastive loss that jointly aligns both the image and masked image embeddings with the text embeddings. The loss for the original image-text pair is defined as:

\begin{equation}
\mathcal{L}_{I} = -\log \frac{e^{\text{s}(E_I, E_T)/\tau}}{\sum_{k=1}^{2N} \mathbb{1}_{[k \neq i]} e^{\text{s}(z_i, z_k)/\tau}}
\end{equation}

where \( \text{s}(u, v) = u^\top v / (\|u\|\|v\|) \) is the cosine similarity, \( E_V \) is the embedding of the i-th image or heatmap-highlighted image, \( E_T \) is the corresponding text embedding, \( N \) is the batch size, \( \tau \) is the temperature parameter, and \( \mathbb{1}_{[k \neq i]} \) is the indicator function equal to 1 iff \( k \neq i \).

Similarly, the loss for the heatmap-highlighted image-text pair is given by:

\begin{equation}
\mathcal{L}_{H} = -\log \frac{e^{\text{s}(E_H, E_T)/\tau}}{\sum_{k=1}^{2N} \mathbb{1}_{[k \neq i]} e^{\text{s}(z_i, z_k)/\tau}}
\end{equation}

The total loss is the average of the two:

\begin{equation}
\mathcal{L}_{\text{SFA}} = \frac{1}{2N} \sum_{(I, T, H) \in \mathcal{D}} \left( \mathcal{L}_{I} + \mathcal{L}_{H} \right)
\end{equation}

By enforcing this dual alignment, SFA ensures that both the global context from the original image and the focused regions from the masked image are jointly considered, leading to improved performance on human-centric tasks. This mechanism effectively guides attention toward individuals in complex scenes (\cref{fig:heatmap_examples}).

\subsubsection{Zero-Shot Inference}

During zero-shot prediction, the goal is to apply the learned function \( f \) to predict the class label \( \hat{c} \) for a new image \( \hat{I} \) without the aid of the heatmap. This is achieved by defining a set of potential class labels \( \mathcal{C} = \{c_1, c_2, \ldots, c_K\} \) and generating a corresponding set of texts \( \mathcal{T} = \{T_1, T_2, \ldots, T_K\} \) through a task-specific sentence template populated with each class label. These texts are then encoded into embeddings by $f$, producing \( \mathcal{Z} = \{E_{T_1}, E_{T_2}, \ldots, E_{T_K}\} \). Concurrently, the embedding of the new image $E_{\hat{I}}$ is obtained. The class label $c$ associated with the text embedding $E_{T_c}$ that has the highest cosine similarity with $E_{\hat{I}}$ is predicted as the class for $\hat{I}$, formalized as:

\begin{equation}
\hat{c} = \argmax_{c \in \mathcal{C}} \left(\text{s}(E_{\hat{I}}, E_{T_c})\right)
\end{equation}

The model performance is measured on an unseen test dataset, which examines the model's ability to generalize to classes not seen during training.






\section{Experiments}
\label{sec:results}

We perform two sets of experiments to optimize our pose descriptions dataset and assess our FocusCLIP framework in zero-shot human classification tasks.

\subsection{Quality evaluation of pose descriptions dataset}

\begin{figure*}[ht]
    \centering
    \scriptsize
    \begin{subfigure}[b]{0.27\linewidth}
        \centering
        \begin{subfigure}[b]{\linewidth}
        \centering
        \includegraphics[width=\linewidth]{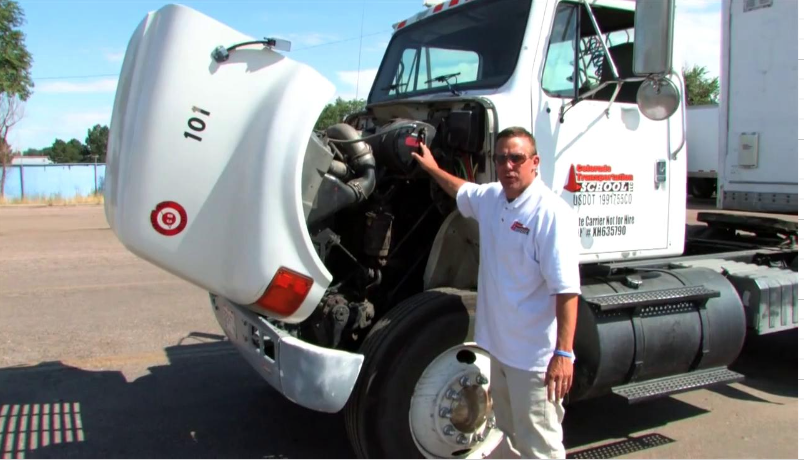}
        \end{subfigure}
        \textbf{loading and unloading a truck}

        \begin{mdframed}[innerleftmargin=4pt, innerrightmargin=4pt, innertopmargin=4pt, innerbottommargin=4pt]
            \scriptsize
            \textbf{GPT-3.5*:} \perfect{There is 1 person in the image }\worst{who is loading and unloading}\neutral{ a truck, }\worst{tying down the load,}\perfect{ standing, }\worst{walking, and carrying a heavy load.}\perfect{ The person is standing }\good{with their legs slightly apart}\neutral{ and their arms extended}\bad{ towards the truck bed.}\perfect{ Their left leg is slightly bent}\worst{ while their right leg is straight. Their left arm is holding a rope to tie down the load while their right arm is holding a heavy box. }\perfect{Their torso is slightly twisted}\worst{ towards the truck bed.}\perfect{ Their head is facing forward.}
        \end{mdframed}
    \end{subfigure}
    \hfill
    \begin{subfigure}[b]{0.34\linewidth}
        \begin{mdframed}[innerleftmargin=4pt, innerrightmargin=4pt, innertopmargin=4pt, innerbottommargin=4pt]
            \scriptsize
            \textbf{GPT-4:} \perfect{There is one person in the image who is working, specifically truck}\worst{ driving, loading and unloading a truck, tying down load, }\perfect{standing,}\worst{ walking and carrying heavy loads. }\perfect{This activity involves a lot of movement and physical exertion. The centrally located person is actively engaged in work with their limbs positioned to carry out tasks. Their right leg seems to be bent at the knee as it's visible from the hip to the knee but the ankle isn't visible suggesting it might be obscured behind an object or positioned towards the back. The left leg appears straighter with both hip and knee visible but like the right leg, its ankle isn't visible either. For their arms; their right arm appears extended outward with a bend at elbow while holding something given that both wrist and elbow are higher than shoulder level. Their left arm also seems to be bent at elbow but unlike right arm, it's likely hanging downwards since its wrist is lower than shoulder level. Their torso leans slightly towards right}\worst{ indicating they might be trying to balance something heavy they're lifting or carrying on their right side.}\perfect{ Lastly for head position; since upper neck and head top keypoints are close together compared to other parts of body suggests they're looking straight ahead}\good{ or slightly downwards which could mean they're focused on task at hand.}
        \end{mdframed}
    \end{subfigure}
    \hfill
    \begin{subfigure}[b]{0.37\linewidth}
        \begin{mdframed}[innerleftmargin=4pt, innerrightmargin=4pt, innertopmargin=4pt, innerbottommargin=4pt]
            \scriptsize
            \textbf{LLaMA-2:} \perfect{There is one person in the image}\worst{ who is occupation, }\neutral{truck}\worst{ driving, loading and unloading truck, tying down load, }\perfect{standing,}\worst{ walking and carrying heavy loads.}\perfect{ The person is standing with their feet shoulder-width apart, with their left foot slightly in front of their right foot.} \neutral{Their left leg is bent at the knee, and their right leg is straight. Their left arm is extended overhead, and their right arm is bent and holding onto something.}\good{ Their torso is straight, and their head is facing forward.}\worst{ The person's left arm is positioned above their right shoulder, and their right arm is positioned below their left shoulder.}\perfect{ Their left leg is positioned in front of their right leg,}\neutral{ and their right leg is positioned behind their left leg. }\worst{The person is wearing a jacket}\good{ and pants, and they have a serious expression on their face.}\perfect{ They are in a standing pose, with their weight evenly distributed on both feet. }\worst{They are holding a rope in their right hand, }\perfect{and their left hand is on their hip. }\bad{The person's body language suggests that they are focused on their task}\worst{ and are exerting effort to tie down the load. Their facial expression indicates that they are concentrating and may be slightly straining from the weight of the load. Overall, the person in the image is engaged in a physically demanding activity that requires strength, endurance, and attention to detail.}\neutral{ They are dressed appropriately for the task, and their body language conveys a sense of determination and focus.}
        \end{mdframed}
    \end{subfigure}
    \caption{\textbf{Qualitative Comparison of LLM Output.} GPT-4 descriptions are more descriptive compared to LLaMA-2. They also contain a higher percentage of \perfect{sentences rated correct} by human evaluators. GPT-4 uses a diverse vocabulary with few repetitions. In comparison, LLaMA-2 output exhibits sentences with repeating patterns. It is also more often \worst{incorrect (i.e., not correlated to the image)}, and the contents of its descriptions are sometimes trivial. We also show the output of GPT-3.5 with an unrefined prompt.}
    \label{fig:samples_mpii}
\end{figure*}

We designed a human evaluation scheme to evaluate the "correctness" of generated text, with results shown in~\cref{fig:caption_alignment}. Each image was annotated with pose descriptions written by three different LLMs, including GPT-3.5, GPT-4, and LLaMA-2. Descriptions comprise a variable number of sentences. We evaluated the sentence-level quality of these descriptions using human feedback. Selecting 100 samples randomly, we split the corresponding 300 descriptions into sentences. Evaluators were shown image-sentence pairs and asked to rank correctness on a five-point scale by comparing the sentence with the corresponding image. The results of this evaluation are illustrated in~\cref{fig:human_eval_results_details}. We also take a closer look at one complete sample with assigned correctness scores by evaluators in~\cref{fig:samples_mpii}. The human evaluation further confirmed GPT-4's superiority in creating contextually relevant and interpretive descriptions, ranking highest among human evaluators. We also used these evaluations to tune our LLM prompt to obtain high-quality descriptions.

\begin{table*}[ht]
    \begin{minipage}{0.63\linewidth}
    \centering
    \scriptsize
    \captionof{table}{\textbf{Quality of pose descriptions} using automated metrics. \textit{*Using shorter response length and prompt without personas or structured syntax.}}
    \label{tab:comparative_evaluation}
    \begin{tabular}{lcccccc}
    \toprule
    & \textbf{Readable} $\uparrow$ & \textbf{Errors} $\downarrow$& \textbf{Diversity} $\uparrow$ & \textbf{Repetition} $\downarrow$ & \textbf{Correlation} $\uparrow$ & \textbf{Correct} $\uparrow$ \\
    \midrule
    {arXiv}
    & 14.47 & 7.75 & 118.73 & 0.16 &  9.54 & - \\
    {GPT-3.5$^*$}
    &  5.61 & 1.32 &  41.85 & 0.40 & 18.07$\dagger$ & 74.8 \\
    \midrule
    {LLaMA-2}
    &  7.52 & 2.45 &  44.37 & 0.57 & 17.28 & 72.0 \\
    {GPT-3.5}
    &  7.63 & 2.79 &  69.23 & 0.36 & 17.46 & 77.6 \\
    {GPT-4}
    & \textbf{10.29} & \textbf{2.01} & \textbf{138.68} & \textbf{0.17} & \textbf{17.97} & \textbf{82.6} \\
    \bottomrule
    \end{tabular}
    \vspace{0.2cm}
    
    \begin{minipage}{0.41\linewidth}
        \centering
        \begin{tikzpicture}
        \begin{axis}[
            scale only axis,
            symbolic x coords={LLaMA-2, GPT-3.5, GPT-4},
            xtick=data,
            ymin=50,
            ymajorgrids=true,
            xticklabel style={anchor=north, font=\scriptsize},
            yticklabel style={font=\scriptsize},
            legend style={at={(0.48,-0.27)}, anchor=north,legend columns=-1,font=\scriptsize},
            grid=none,
            width=0.9\linewidth,
            height=1.2cm,
        ]
        
        \addplot[
            color=cyan,
            mark=o,
            thick,
            mark options={solid}
            ] coordinates {
            (LLaMA-2,86.4)
            (GPT-3.5,87.3)
            (GPT-4,89.85)
        };
        
        \addplot[
            color=violet,
            mark=square,
            dashed,
            mark options={solid}
            ] coordinates {
            (LLaMA-2,72)
            (GPT-3.5,77.6)
            (GPT-4,82.6)
        };
        
        \legend{CLIP Score, Human Score}
        
        \end{axis}
        \end{tikzpicture}
        \captionof{figure}{GPT-4's descriptions correlate more with images according to CLIP Score and human feedback.}
        \label{fig:caption_alignment}
    \end{minipage}
    \hfill
    \begin{minipage}{0.57\linewidth}
      \centering
        \begin{tikzpicture}
        \begin{axis}[
            scale only axis,
            cycle list={
              {blue,mark=*,dashed},
              {orange,mark=square*,dotted,very thick},
              {green,mark=triangle*},
            },
            symbolic x coords={LLaMA-2, GPT-3.5, GPT-4},
            xtick=data,
            ymin=0.2, ymax=1.2,
            ymajorgrids=true,
            xticklabel style={anchor=north,font=\scriptsize},
            yticklabel style={font=\scriptsize},
            legend style={at={(0.46,-0.27)}, anchor=north,legend columns=-1,font=\scriptsize},
            width=0.87\linewidth,
            height=1.2cm
        ]
        
        \addplot+ [error bars/.cd, y dir=both, y explicit] coordinates {
            (LLaMA-2,   0.5197)
            (GPT-3.5,   0.5273)
            (GPT-4,     0.7104)
        };
        
        \addplot+ [error bars/.cd, y dir=both, y explicit] coordinates {
            (LLaMA-2,   0.3737)
            (GPT-3.5,   0.5831)
            (GPT-4,     1.1680)
        };
        
        \addplot+ [error bars/.cd, y dir=both, y explicit] coordinates {
            (LLaMA-2,   0.2807)
            (GPT-3.5,   0.4444)
            (GPT-4,     0.9412)
        };
        
        \addplot[
            color=black,
            mark=none,
            dashed,
            thick
            ] coordinates {
            (LLaMA-2,   1.0)
            (GPT-3.5,   1.0)
            (GPT-4,     1.0)
        };
        
        \legend{Readability, Diversity, Repetition (Inv)}
        \end{axis}
        \end{tikzpicture}
        \captionof{figure}{GPT-4 surpassed human-written text in diversity and readability while maintaining a low repetition rate.}
        \label{fig:comparative_evaluation_chart}
    \end{minipage}
    \end{minipage}
    \hfill
    \begin{minipage}{0.35\linewidth}
    \centering
    \begin{tikzpicture}
      \begin{axis}[
          scale only axis,
          ybar,
          bar width=.2cm,
          width=0.92\linewidth,
          height=1.95cm,
          enlarge x limits=0.2,
          legend style={
              at={(0.5,-0.4)},
              anchor=north,
              legend columns=3,
              legend cell align={left},
              font=\scriptsize
          },
          symbolic x coords={LLaMA-2, GPT-3.5, GPT-4},
          xtick=data,
          xticklabel style={
              anchor=north,
              font=\scriptsize
          },
          yticklabel style={font=\scriptsize},
          axis background/.style={fill=ontertiarycontainer},
      ]
    
      \addplot+[score1, fill=score1] coordinates {(GPT-3.5,10.6) (LLaMA-2,15.5) (GPT-4,8.4)};
      \addplot+[score2, fill=score2] coordinates {(GPT-3.5,7.5) (LLaMA-2,10.6) (GPT-4,5.5)};
      \addplot+[score3, fill=score3] coordinates {(GPT-3.5,11.9) (LLaMA-2,9.3) (GPT-4,5.7)};
      \addplot+[score4, fill=score4] coordinates {(GPT-3.5,22.9) (LLaMA-2,28.0) (GPT-4,25.1)};
      \addplot+[score5, fill=score5] coordinates {(GPT-3.5,47.1) (LLaMA-2,36.6) (GPT-4,55.3)};
    
      \legend{Wrong, Partly Wrong, Neutral, Mostly Correct, Correct}
      \end{axis}
      \end{tikzpicture}
    \captionof{figure}{\textbf{Human evaluation of LLM outputs.} The image-sentence pairs were given to human evaluators who rated them on a five-step scale from \worst{completely wrong (1)} to \perfect{perfect (5)}, with three intermediate steps \bad{ 2 } \neutral{ 3 } \good{ 4 }. The evaluators marked the largest proportion of GPT-4 outputs as "correct", with only $\leq 15\%$ of GPT-4 sentences partially or completely incorrect.}
    \label{fig:human_eval_results_details}
    \end{minipage}
\end{table*}

We also compared LLM-written descriptions with a baseline of human-written text--specifically scientific abstracts from \texttt{arXiv}. We used several metrics to assess the quality of the descriptions, including Flesch-Kincaid Grade Level for readability level, number of grammatical errors for text integrity, MTLD for lexical diversity, and a 3-gram repetition score for text uniqueness. Moreover, we used the CLIP Score metric with OpenAI's CLIP-ViT-L/14 for image-description correlation.~\cref{tab:comparative_evaluation} shows that GPT-4 generated descriptions offer greater readability, fewer grammatical errors, enhanced linguistic diversity, and lower repetition than other models. This suggests GPT-4's proficient use of an extensive vocabulary and nuanced understanding of depicted activities, surpassing even human-generated text in diversity (\cref{fig:comparative_evaluation_chart}). GPT-4 provides valuable insights in addition to limb locations and orientations. For instance, when describing a person playing football with one leg straight and the other bent (\cref{tab:data-samples}), only GPT-4 reasoned that the person's weight must be more on the straight leg.

\subsection{Evaluation on human classification tasks}

In this set of experiments, we evaluate the zero-shot classification abilities of FocusCLIP, which was trained on MPII Human Pose dataset~\cite{andriluka14cvpr} images and our pose descriptions. We apply the model to three human-centric tasks, using five previously unseen datasets. Our main baseline is vanilla CLIP training~\cite{radford2021learning} from scratch on the same image-text pairs as FocusCLIP for a fair comparison. We also report a lower bound for accuracy by simulating a random guess averaged over three runs.

\noindent \textbf{Implementation details.} \hspace{2pt}  We initialize the visual encoder with ImageNet pre-trained weights and freeze it, in line with recommendations by~\cite{sun2023eva, zhai2022lit}. Our models are trained using a contrastive loss, with a fixed temperature value of 0.5, consistent with the approach used in SimCLR~\cite{chen2020simple}. We utilize a Stochastic Gradient Descent (SGD) optimizer, with a learning rate of 0.001 and momentum of 0.9, and train all models for 64 epochs. This process takes approximately 12 hours on an A100-40GB GPU. For all our experiments, we use ViT-B/16~\cite{dosovitskiy2020image} for visual and ROI encoders and BERT-Base~\cite{devlin2018bert} for the textual encoder.

\begin{table}[ht]
    \centering
    \scriptsize
    \caption{FocusCLIP improves CLIP's top-k accuracy by 8.61\% for three zero-shot human classification tasks covering five unique datasets. \textit{Best values are \textbf{bold} and our method is {\sethlcolor{lightgray}\hl{highlighted}}. Average is reported across task categories}}
    \begin{tabular}{llrcccgcc}
    \toprule
    \textbf{Task} & \textbf{Dataset} & \textbf{C} & \textbf{k} & \textbf{Chance} & \textbf{CLIP} & \textbf{Ours} \\
    \cmidrule{1-4} \cmidrule(l){5-6} \cmidrule(l){7-7} \cmidrule(l){8-8}
    \multirow{1}{*}{\makecell[l]{{Activity}}}
     & {Stanford40}     &  40 & 3 &  8.24 &   6.49 & \textbf{10.47} \\
    \cmidrule{1-4} \cmidrule(l){5-6} \cmidrule(l){7-7} \cmidrule(l){8-8}
    \multirow{4}{*}{\makecell[l]{{Age}}}
     & {Emotic}      &   3 & 1 & 33.54 & 37.56 & \textbf{41.80} \\
     & {LAGENDA-Body}     &   3 & 1 & 33.46 & 39.48 & \textbf{59.44} \\
     & {LAGENDA-Face}     &   3 & 1 & 33.13 & 44.56 & \textbf{71.41} \\
     & {UTKFace}        &   5 & 1 & 20.86 & 27.02 & \textbf{35.13} \\
    \cmidrule{1-4} \cmidrule(l){5-6} \cmidrule(l){7-7} \cmidrule(l){8-8}
    \multirow{2}{*}{\makecell[l]{{Emotion}}}
     & {Emotic}      &  26 & 3 & 11.54 & 10.39 & \textbf{13.73} \\
     & {FER+}        &   8 & 3 & 36.79 & 52.56 & \textbf{63.35} \\
    \cmidrule(l){5-8}
    \multicolumn{4}{r}{\textbf{Task-wise Mean}} & 20.88 & 25.04 & \textbf{33.65} \\
    \bottomrule
    \end{tabular}
    \label{tab:results_zs}
\end{table}

\cref{tab:results_zs} presents our quantitative results. We report the top-k accuracy for image-based activity classification, age classification, and emotion recognition. For datasets with numerical age labels, we categorize the age into groups such as adult, teenager, kid, etc., to transform it into a classification task. FocusCLIP outperforms the baseline CLIP by 3.98\% with 10.47\% accuracy on the Stanford40~\cite{yao2011human} dataset for activity recognition. Similarly, significant improvements are observed across various age classification datasets, including Emotic~\cite{kosti2017emotic} and LAGENDA-Body~\cite{mivolo2023}, which contain full body images, and LAGENDA-Face~\cite{mivolo2023} and UTKFace~\cite{zhang2017age}, which contain cropped facial images. The improvements range from 4.24\% to 26.85\%. In the emotion recognition task, there is a similar trend with 3.34\% improvement for Emotic~\cite{kosti2017emotic} and 10.79\% improvement for FER+~\cite{BarsoumICMI2016}. The improvements in zero-shot classification performance across diverse human-centric tasks illustrate FocusCLIP's enhanced ability to understand and interpret human-centric features more effectively than the CLIP baseline. Notably, the improvements in age classification across various datasets highlight its capacity to grasp implicit human attributes, and significant gains in emotion recognition tasks emphasize its proficiency in identifying emotional cues from both body language (Emotic) and facial expressions (FER+).

Furthermore, FocusCLIP's performance in the activity recognition task underscores its capability to contextualize human actions within a scene, suggesting that the model can extract meaningful information from both the foreground and background. As we show in our ablation studies, heatmaps that additionally highlight important scene elements further improve performance for this task. This supports our claim that heatmaps are a versatile and effective way to guide model focus.

These results suggest that combining pose descriptions with heatmap-based guidance in FocusCLIP effectively directs the model's focus toward relevant features within an image, enabling it to make more accurate predictions. This highlights the potential of specialized training for models that not only perform well in narrow, task-specific benchmarks but also exhibit a broader understanding of human-centric concepts in a more generalized, zero-shot context.

\begin{table*}[ht]
    \begin{minipage}{0.68\linewidth}
    \centering
    \scriptsize
    \captionof{table}{\textbf{Impact of SFA}. We randomly select images from the test datasets and show the top three predicted classes. The correct label is highlighted in \textcolor{primarycontainer}{\sethlcolor{primary}\hl{ bright purple }}. A prediction is correct if the purple bar appears at the top. Without SFA, the model either fails to predict the correct class \sethlcolor{errorcontainer}\hl{(highlighted in red)} or has lower confidence}
    \setlength{\tabcolsep}{3pt} 
    \begin{tabular}{ccccccc}
        \toprule
        & \multicolumn{2}{c}{\textbf{Activity Recognition}} &
        \multicolumn{2}{c}{\textbf{Age Classification}} &
        \multicolumn{2}{c}{\textbf{Emotion Recognition}} \\
        \cmidrule(l){2-3} 
        \cmidrule(l){4-5}
        \cmidrule(l){6-7}
        \rotatebox{90}{no SFA} &
        \multirow{2}{*}[1.8em]{\includegraphics[width=1.25cm]{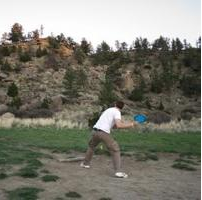}} &
        \cellcolor{errorcontainer}
        \begin{tikzpicture}
            \begin{axis}[
                xbar,
                xmin=20,
                yticklabels=\empty,
                bar width=5pt,
                grid=major,
                width=2.3cm,
                height=3cm,
                nodes near coords,
                nodes near coords align={west},
                nodes near coords style={font=\scriptsize},
                point meta=explicit symbolic, 
                font=\scriptsize,
                axis lines=none
            ]
            \addplot[fill=secondary, draw=secondary] coordinates {(45.1,2) [cutting trees]};
            \addplot[fill=primary, draw=primary] coordinates {(39.8,1) [throwing frisby]};
            \addplot[fill=secondary, draw=secondary] coordinates {(37.2,0) [walking the dog]};
            \end{axis}
        \end{tikzpicture} &
        \multirow{2}{*}[1.8em]{\includegraphics[width=1.25cm]{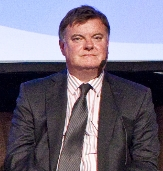}} &
        \cellcolor{errorcontainer}
        \begin{tikzpicture}
            \begin{axis}[
                xbar,
                xmin=20,
                yticklabels=\empty,
                bar width=5pt,
                grid=major,
                width=2.3cm,
                height=3cm,
                nodes near coords,
                nodes near coords align={west},
                nodes near coords style={font=\scriptsize},
                point meta=explicit symbolic, 
                font=\scriptsize,
                axis lines=none
            ]
            \addplot[fill=secondary, draw=secondary] coordinates {(34.17,2) [teenager]};
            \addplot[fill=primary, draw=primary] coordinates {(33.76,1) [adult]};
            \addplot[fill=secondary, draw=secondary] coordinates {(32.07,0) [kid]};
            \end{axis}
        \end{tikzpicture} &
        \multirow{2}{*}[2.2em]{\includegraphics[width=1.25cm]{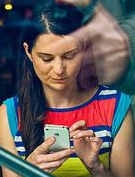}} &
        \begin{tikzpicture}
            \begin{axis}[
                xbar,
                xmin=20,
                yticklabels=\empty,
                bar width=5pt,
                grid=major,
                width=2.3cm,
                height=3cm,
                nodes near coords,
                nodes near coords align={west},
                nodes near coords style={font=\scriptsize},
                point meta=explicit symbolic, 
                font=\scriptsize,
                axis lines=none
            ]
            \addplot[fill=primary, draw=primary] coordinates {(39.6,2) [disconnection]};
            \addplot[fill=secondary, draw=secondary] coordinates {(39.5,1) [fear]};
            \addplot[fill=secondary, draw=secondary] coordinates {(39.4,0) [pain]};
            \end{axis}
        \end{tikzpicture}  \\
        \cmidrule(l){1-1} 
        \cmidrule(l){3-3} 
        \cmidrule(l){5-5}
        \cmidrule(l){7-7}
        \rotatebox{90}{SFA} & & 
        \begin{tikzpicture}
            \begin{axis}[
                xbar,
                xmin=20,
                yticklabels=\empty,
                bar width=5pt,
                grid=major,
                width=2.3cm,
                height=3cm,
                nodes near coords,
                nodes near coords align={west},
                nodes near coords style={font=\scriptsize},
                point meta=explicit symbolic, 
                font=\scriptsize,
                axis lines=none
            ]
            \addplot[fill=primary, draw=primary] coordinates {(47.3,2) [throwing frisby]};
            \addplot[fill=secondary, draw=secondary] coordinates {(45.9,1) [cutting trees]};
            \addplot[fill=secondary,draw=secondary] coordinates {(42.7,0) [walking the dog]};
            \end{axis}
        \end{tikzpicture} & &
        \begin{tikzpicture}
            \begin{axis}[
                xbar,
                xmin=20,
                yticklabels=\empty,
                bar width=5pt,
                grid=major,
                width=2.3cm,
                height=3cm,
                nodes near coords,
                nodes near coords align={west},
                nodes near coords style={font=\scriptsize},
                point meta=explicit symbolic, 
                font=\scriptsize,
                axis lines=none
            ]
            \addplot[fill=primary, draw=primary] coordinates {(34.48,2) [adult]};
            \addplot[fill=secondary, draw=secondary] coordinates {(33.75,1) [teenager]};
            \addplot[fill=secondary, draw=secondary] coordinates {(31.76,0) [kid]};
            \end{axis}
        \end{tikzpicture} & &
        \begin{tikzpicture}
            \begin{axis}[
                xbar,
                xmin=20,
                yticklabels=\empty,
                bar width=5pt,
                grid=major,
                width=2.3cm,
                height=3cm,
                nodes near coords,
                nodes near coords align={west},
                nodes near coords style={font=\scriptsize},
                point meta=explicit symbolic, 
                font=\scriptsize,
                axis lines=none
            ]
            \addplot[fill=primary, draw=primary] coordinates {(43.34,2) [disconnection]};
            \addplot[fill=secondary, draw=secondary] coordinates {(40.80,1) [confusion]};
            \addplot[fill=secondary, draw=secondary] coordinates {(40.00,0) [sympathy]};
            \end{axis}
        \end{tikzpicture} \\
        \midrule
        \rotatebox{90}{no SFA} &

        \multirow{2}{*}[1.8em]{\includegraphics[width=1.25cm]{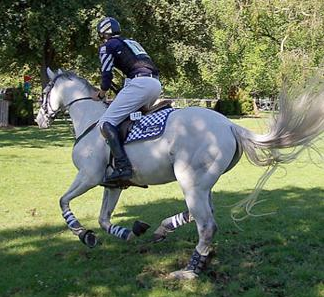}} &
        \cellcolor{errorcontainer}
        \begin{tikzpicture}
            \begin{axis}[
                xbar,
                xmin=20,
                yticklabels=\empty,
                bar width=5pt,
                grid=major,
                width=2.3cm,
                height=3cm,
                nodes near coords,
                nodes near coords align={west},
                nodes near coords style={font=\scriptsize},
                point meta=explicit symbolic, 
                font=\scriptsize,
                axis lines=none
            ]
            \addplot[fill=secondary, draw=secondary] coordinates {(53.2,2) [feeding a horse]};
            \addplot[fill=primary, draw=primary] coordinates {(51.5,1) [riding a horse]};
            \addplot[fill=secondary, draw=secondary] coordinates {(30.3,0) [riding a bike]};
            \end{axis}
        \end{tikzpicture} &

        \multirow{2}{*}[1.8em]{\includegraphics[width=1.25cm]{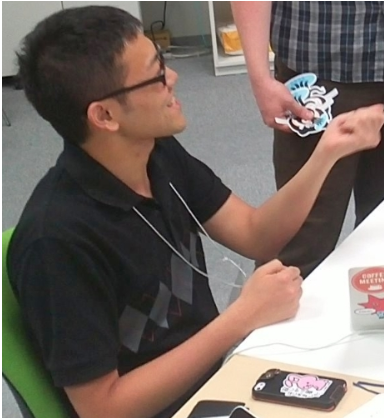}} &
        \begin{tikzpicture}
            \begin{axis}[
                xbar,
                xmin=20,
                yticklabels=\empty,
                bar width=5pt,
                grid=major,
                width=2.3cm,
                height=3cm,
                nodes near coords,
                nodes near coords align={west},
                nodes near coords style={font=\scriptsize},
                point meta=explicit symbolic, 
                font=\scriptsize,
                axis lines=none
            ]
            \addplot[fill=primary, draw=primary] coordinates {(34.30,2) [adult]};
            \addplot[fill=secondary, draw=secondary] coordinates {(33.57,1) [teenager]};
            \addplot[fill=secondary, draw=secondary] coordinates {(32.13,0) [kid]};
            \end{axis}
        \end{tikzpicture} &

        \multirow{2}{*}[2.2em]{\includegraphics[width=1.25cm]{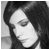}} &
        \cellcolor{errorcontainer}
        \begin{tikzpicture}
            \begin{axis}[
                xbar,
                xmin=20,
                yticklabels=\empty,
                bar width=5pt,
                grid=major,
                width=2.3cm,
                height=3cm,
                nodes near coords,
                nodes near coords align={west},
                nodes near coords style={font=\scriptsize},
                point meta=explicit symbolic, 
                font=\scriptsize,
                axis lines=none
            ]
            \addplot[fill=secondary, draw=secondary] coordinates {(31.32,2) [happiness]};
            \addplot[fill=primary, draw=primary] coordinates {(30.58,1) [sadness]};
            \addplot[fill=secondary, draw=secondary] coordinates {(29.12,0) [disgust]};
            \end{axis}
        \end{tikzpicture}  \\
        \cmidrule(l){1-1} 
        \cmidrule(l){3-3} 
        \cmidrule(l){5-5}
        \cmidrule(l){7-7}
        \rotatebox{90}{SFA} & & 

        \begin{tikzpicture}
            \begin{axis}[
                xbar,
                xmin=20,
                yticklabels=\empty,
                bar width=5pt,
                grid=major,
                width=2.3cm,
                height=3cm,
                nodes near coords,
                nodes near coords align={west},
                nodes near coords style={font=\scriptsize},
                point meta=explicit symbolic, 
                font=\scriptsize,
                axis lines=none
            ]
            \addplot[fill=primary, draw=primary] coordinates {(48.2,2) [riding a horse]};
            \addplot[fill=secondary, draw=secondary] coordinates {(47.1,1) [feeding a horse]};
            \addplot[fill=secondary,draw=secondary] coordinates {(37.1,0) [walking the dog]};
            \end{axis}
        \end{tikzpicture} & &
        \begin{tikzpicture}
            \begin{axis}[
                xbar,
                xmin=20,
                yticklabels=\empty,
                bar width=5pt,
                grid=major,
                width=2.3cm,
                height=3cm,
                nodes near coords,
                nodes near coords align={west},
                nodes near coords style={font=\scriptsize},
                point meta=explicit symbolic, 
                font=\scriptsize,
                axis lines=none
            ]
            \addplot[fill=primary, draw=primary] coordinates {(35.16,2) [adult]};
            \addplot[fill=secondary, draw=secondary] coordinates {(33.17,1) [teenager]};
            \addplot[fill=secondary, draw=secondary] coordinates {(31.67,0) [kid]};
            \end{axis}
        \end{tikzpicture} & &
        \begin{tikzpicture}
            \begin{axis}[
                xbar,
                xmin=20,
                yticklabels=\empty,
                bar width=5pt,
                grid=major,
                width=2.3cm,
                height=3cm,
                nodes near coords,
                nodes near coords align={west},
                nodes near coords style={font=\scriptsize},
                point meta=explicit symbolic, 
                font=\scriptsize,
                axis lines=none
            ]
            \addplot[fill=primary, draw=primary] coordinates {(31.10,2) [sadness]};
            \addplot[fill=secondary, draw=secondary] coordinates {(29.58,1) [happiness]};
            \addplot[fill=secondary, draw=secondary] coordinates {(29.24,0) [neutral]};
            \end{axis}
        \end{tikzpicture} \\
        \bottomrule
    \end{tabular}
    \label{tab:qualitative_results}
    \end{minipage}
    \hfill
    \begin{minipage}{0.3\linewidth}
        \centering
        \scriptsize
        \captionof{table}{\textbf{Our structured prompt} creates better pose descriptions, improving downstream performance}
        \label{tab:ablation_prompts}
        \vspace{-0.2cm}
        \begin{tabularx}{\linewidth}{Xgc}
        \toprule
         \textbf{Task} & \textbf{Structured} & \textbf{Plain}\\
         \midrule
         Activity  & \textbf{10.47} &  6.18 \\
         Age       & \textbf{51.94} & 45.40 \\
         Emotion   & 38.54 & \textbf{40.32} \\
         \midrule
         Mean      & \textbf{33.65} & 30.63 \\
         \bottomrule
        \end{tabularx}
        \vspace{0.2cm}

        \begin{tikzpicture}
            \begin{axis}[
                scale only axis,
                ybar,
                xtick=data,
                xticklabels={Readability, Diversity, Repetition, Correlation},
                enlarge x limits=0.15,
                legend style={at={(0.5,-0.5)}, anchor=north,legend columns=-1, font=\scriptsize},
                xticklabel style={font=\scriptsize},
                yticklabel style={font=\scriptsize},
                bar width=8pt,
                ymin=3, ytick=\empty,
                grid=minor,
                width=0.95\linewidth,
                height=1cm,
            ]
    
            \addplot[
                fill=tertiary,
                draw=tertiary
            ] coordinates {
                (0,5.61) 
                (1,4.185)
                (2,4.0)
                (3,7.48)
            };
    
            \addplot[
                fill=primary,
                draw=primary
            ] coordinates {
                (0,7.63)
                (1,6.923)
                (2,3.6)
                (3,7.76)
            };
        
            \legend{Unrefined Prompt, Our Prompt}
            \end{axis}
        \end{tikzpicture}
        \vspace{-0.2cm}
        \captionof{figure}{\textbf{Using contextual cues and personas} in our prompt enhances vocabulary diversity, reduces repetition, increases readability, and improves image correlation to generated text}
        \label{fig:ablation_prompt_unrefined}
    \end{minipage}
\end{table*}

\begin{table*}[ht]
    \begin{minipage}{0.37\linewidth}
        \centering
        \scriptsize
        \captionof{table}{\textbf{FocusCLIP component ablations.} Heatmap-masked images (MIX) and sharing weights (SE) have the most impact}
        \vspace{-0.2cm}
        \setlength{\tabcolsep}{3pt} 
        \begin{tabular}{cccc|ccc|c}
        \toprule
         ROI & $\mathcal{L}_{R,T}$ & \textbf{SE} & \textbf{MIX} & \textbf{Activity} & \textbf{Age} & \textbf{Emotion} & \textbf{Mean} \\
         \midrule
                & \      &        &        &  6.49 & 37.16 & 31.48 & 25.04 \\
         \cmark &        & \cmark & \cmark &  3.47 & 49.28 & \textbf{39.68} & 30.81 \\
         \cmark & \cmark &        & \cmark &  4.66 & 40.96 & 26.68 & 24.10 \\
         \cmark & \cmark & \cmark &        &  8.06 & 46.32 & 21.52 & 25.30 \\
         \rowcolor[gray]{0.8}
         \cmark & \cmark & \cmark & \cmark & \textbf{10.47} & \textbf{51.94} & 38.54 & \textbf{33.65} \\
         \bottomrule
        \end{tabular}
        \label{tab:ablation_results}
    \end{minipage}
    \hfill
    \begin{minipage}{0.33\linewidth}
        \centering
        \scriptsize
        \captionof{table}{\textbf{Ablating over Heatmaps:} Evaluation of keypoint-based, DINO, TCL, and bounding-box heatmaps on various tasks}
        \vspace{-0.2cm}
        \setlength{\tabcolsep}{3pt} 
        \begin{tabular}{lgcccc}
        \toprule
         \textbf{Task} & \textbf{Keypoint} & \textbf{Box} & \textbf{DINO} & \textbf{TCL} & \textbf{None} \\
        \midrule
        Activity & 10.47 &  3.74 & \textbf{13.38} &  3.49 &  6.49 \\
        Age      & \textbf{51.94} & 41.16 & 39.54 & 46.86 & 37.16 \\
        Emotion  & \textbf{38.54} & 32.20 & 35.90 & 35.87 & 31.48 \\
        \midrule
        Mean     & \textbf{33.65} & 25.70 & 29.61 & 28.74 & 25.04 \\
        \bottomrule
        \end{tabular}
        \label{tab:ablation_heatmaps}
    \end{minipage}
    \hfill
    \begin{minipage}{0.28\linewidth}
        \centering
        \scriptsize
        \captionof{table}{\textbf{Ablating over LLMs}: Downstream performance comparison using different LLMs}
        \vspace{-0.2cm}
        \setlength{\tabcolsep}{3pt} 
        \begin{tabular}{lgcc}
        \toprule
        \textbf{Task} & \textbf{GPT-4} & \textbf{GPT-3.5} & \textbf{LLaMA-2}\\
         \midrule
         {Activity} & \textbf{10.47} &  7.97 &  4.88 \\
         {Age}      & \textbf{51.94} & 36.41 & 40.69 \\
         {Emotion}  & 38.54 & 38.71 & \textbf{39.82} \\
         \midrule
         {Mean} & \textbf{33.65} & 27.70 & 28.46 \\
         \bottomrule
        \end{tabular}
        \label{tab:ablation_llms}
    \end{minipage}
\end{table*}

\section{Ablations studies}
\label{sec:ablations}

The major components of FocusCLIP include the ROI encoder and dual contrastive loss. Furthermore, we multiply the heatmap with the original image to create a highlighted image before feeding it into the ROI encoder, and we share the weights between both visual encoders. We analyze the value of each of these four components in Table~\cref{tab:ablation_results}. By removing each component separately and observing the impact on performance, we determine that all components work together to boost performance. The ROI encoder with heatmap input and the multiplication operation to highlight image regions (MIX) played important roles and notably impacted performance (rows 1 and 4). However, the most significant decrease in performance was observed when the two visual encoders did not share weights (row 3). This was due to the increased complexity of the learning objective and the doubling of model parameters, resulting in overfitting on our small training data and inferior zero-shot performance on unseen datasets. The additional loss function (row 2) also contributes to the overall performance of our network (row 5).

\noindent \textbf{Impact of the ROI encoder.} \hspace{2pt} The primary contribution of our work is the heatmap-based attention provided by the ROI encoder. The second contribution is the pose descriptions dataset. As shown in Table~\cref{tab:ablation_results}, when the model is trained with our pose descriptions, it achieves a task-wise mean accuracy of 25.04\% (row 1). This is 4.16\% above the random baseline at 20.88\%, demonstrating the impact of our text dataset. When the person heatmaps are provided as an auxiliary input, the task-wise mean improves to 33.65\% (row 5), an additional 8.61\%. This confirms our initial hypothesis that heatmaps can provide subject-focused supervision, allowing FocusCLIP to learn feature representations better suited for the intended downstream tasks.

\noindent \textbf{Ablations over Prompting Method:} \hspace{2pt} \cref{fig:ablation_prompt_unrefined} compares the statistics of responses generated by two different prompts: the one we proposed, which includes persona and multiple contextual cues, and an unrefined prompt that lacks these features. Quantitative ablations in~\cref{tab:ablation_prompts} reveal that structured prompts, which utilize function-like syntax, yield a higher average performance across various tasks than plain prompts. The results show that our proposed prompt leads to higher-quality responses regarding language and correlation with the images. We present detailed qualitative results in Appendix B, illustrating how adding activity labels and persons enhances the contextual richness of responses and complements the keypoint data. Similarly, our human evaluations validate the hypothesis that embedding image-specific attributes and other contextual data in the prompt results in more precise generated descriptions. This can be explained by multiple contextual cues and image attributes that we add to the prompt. It reduces the expectations from LLM to extrapolate new information by instead asking it to parse structured data, mitigating hallucinations. This emphasizes the significance of prompt engineering when using LLMs as annotators.

\noindent \textbf{Impact of heatmap quality.} \hspace{2pt} We compare three heatmap sources: Gaussian ellipses within object bounding boxes, self-attention maps from a DINO model~\cite{caron2021emerging}, and zero-shot heatmaps generated from TCL~\cite{cha2023learning}. Bounding box-based heatmaps, while less annotation-dependent, lack the shape detail provided by keypoint-based annotations. DINO-derived maps eliminate manual annotation but offer limited control over shape or image regions. TCL heatmaps, generated through textual description, provide flexibility but have low resolution and contrast. As shown in \cref{tab:ablation_heatmaps}, including any heatmap consistently enhances performance over a CLIP-only baseline. Our original fine-grained, keypoint-based heatmaps excel, particularly in age and emotion recognition, achieving the highest average performance of 33.65. DINO-based heatmaps demonstrate a unique advantage in activity recognition, scoring 13.38, as they effectively highlight multiple significant regions within the scene, not just the subject. This contributes to their overall average score of 29.61. TCL-based heatmaps also show competitive performance with an average score of 28.74. However, bounding-box-based heatmaps offer the least improvement, with an average score of 25.70, reiterating their limitations in capturing object intricacies. Notably, DINO and TCL-based heatmaps surpass the no-heatmap baseline without manual annotation. This demonstrates the ability of FocusCLIP to benefit from heatmaps.

\noindent \textbf{Impact of LLM choice on performance.} \hspace{2pt} We examine how different LLMs influence zero-shot classification performance using our single-shot prompting method, as summarized in~\cref{tab:ablation_llms}. GPT-4-generated captions enable our FocusCLIP model to achieve the best performance on the MPII dataset, closely followed by GPT-3.5. Conversely, captions generated by LLaMA-2 lag, likely due to its fewer parameters hindering the effective parsing of our JSON-formatted prompts. These observations are consistent with quality evaluations in~\cref{fig:human_eval_results_details}.

\section{Conclusion}
\label{sec:conclusion}

This paper introduced the MPII Pose Descriptions dataset, a novel contribution that combines annotated human pose images with detailed, context-aware captions generated by LLMs. The dataset enriches human-centric tasks such as keypoint estimation and activity recognition by providing semantically rich descriptions of human poses. We also presented FocusCLIP, a modification of the CLIP framework, which incorporates Subject-Focused Attention to enhance the model's understanding of humans. Combining fine-grained pose descriptions and focused attention mechanisms offers a practical and scalable solution to improve performance across various human-centered visual tasks. Our work establishes a foundation for future research, especially in integrating more sophisticated attention-based fusion strategies between multiple visual inputs and exploring the use of the dataset for tasks beyond zero-shot classification, such as fine-grained pose analysis and text-enhanced keypoint estimation.

\noindent \textbf{Limitations} While our dataset holds significant potential, there are important considerations regarding the automatic generation of pose descriptions. Although we ask LLMs to use gender-neutral language to reduce activity-related bias, the dataset may still inadvertently reflect stereotypes inherent in the LLMs. Furthermore, the correctness of the generated descriptions was validated on only a small subset of the data, which may lead to inaccuracies or inconsistencies. Future iterations can explore improved fusion techniques and bias mitigation strategies. Despite these challenges, the dataset and model contribute valuable resources to advancing human motion and activity understanding research.

{\small
\bibliographystyle{ieee_fullname}
\bibliography{main}
}

\clearpage
\appendix
\appendix
\setcounter{page}{1}
\setcounter{figure}{0}
\setcounter{table}{0}

\section{LLM Prompt}
\label{sec:llm_prompt_mpii}

The section describes the prompt structure, its rationale, and design process. The specific prompt used for data generation is as follows:

\begin{quote}
\textit{You are an expert human activity and pose analyzer with deep understanding of MPII Human Pose dataset, which has 16 keypoints in order: 0 - right ankle, 1 - right knee, 2 - right hip, 3 - left hip, 4 - left knee, 5 - left ankle, 6 - pelvis, 7 - thorax, 8 - upper neck, 9 - head top, 10 - right wrist, 11 - right elbow, 12 - right shoulder, 13 - left shoulder, 14 - left elbow, 15 - left wrist. Given a set of 2D keypoint coordinates from MPII dataset as (x,y) with -1 for invisible joints, you will precisely describe body poses in terms of relative limb locations. Your descriptions will follow this template: ``There are [num2word(\$count)] people in image who are [getVerb(\$activity) parseName(\$activity)]. [General attributes describing \$activity in keypoints context.]'' For each person in image: ``The [parseLocation(\$center,\$scale)] person is [predictStateFromContext()] with their [limb]...'' For each limb (left leg, right leg, left arm, right arm, torso, head): ``[Describe how these limbs are positioned relative to other limbs, bend angles, and other similar pose information.]'' Use concise, precise, and gender-neutral language.}
\end{quote}

\subsection{Design Decisions}
\label{sec:prompt-engineering}

We used insights from contemporary prompt engineering works~\cite{white2023prompt,liu2023jailbreaking,moller2023prompt} to design a prompt that can describe humans in images using an LLM. In particular, our generic template prompt relies on persona specification~\cite{moller2023prompt} and the ability of LLMs to parse structured data~\cite{liu2023lost, guo2023gpt4graph}.

The opening sentence establishes the LLM \textit{role}:

\begin{quote}
\textit{You are an experienced \{role\}, with a deep understanding of \{dataset name\} dataset, which has \{dataset description\}.}
\end{quote}

This is inspired from the \textit{persona patterns} in~\cite{white2023prompt,sivarajkumar2023empirical,sisaengsuwanchai2023does}. In our case, the \textit{role} is ``human pose estimation expert'', \textit{dataset name} is ``MPII Human Pose'', and the \textit{description} includes an explanation of available keypoints and other annotations. Including this effectively morphs the LLM into a knowledgeable entity about the dataset. Inspired by the experiments in~\cite{moller2023prompt}, we interface this \textit{persona specification} with LLM system prompt to further enhance context-aware responsiveness.

Next, we define the \textit{required content} in the generated descriptions, including a \textit{captioning objective} for additional context. 

\begin{quote}
\textit{Given the following annotations of an image from the \{dataset name\} dataset, describe the \{target\} in the image in terms of \{required content\}, and any other discriminatory attributes necessary for \{captioning objective\}:}
\end{quote}

This is followed by the \textit{raw image-specific attributes} from the dataset annotations as key-value pairs. We use the activity label, people count, keypoint locations, visibility, center, and scale for each annotated person. This is our main difference from previous works~\cite{naeem2023i2mvformer,moller2023prompt} who harness the few-shot learning capabilities of LLMs to generate data, providing a handful of examples and prompting LLMs to synthesize analogous data. In our approach, we instead leverage the well-documented ability of LLMs to parse structured data~\cite{liu2023lost, guo2023gpt4graph}.

The next section of our prompt defines the required \textit{response format}. In our experiments, we ask the LLM to fill in a semi-standard template: \textit{Your response should \{response format\}}

\begin{quote}
\textit{``There are [num2word(\$count)] people in image who are [getVerb(\$activity) parseName(\$activity)]. [General attributes describing \$activity in keypoints context.]'' For each person in image: ``The [parseLocation(\$center,\$scale)] person is [predictStateFromContext()] with their [limb]...''}
\end{quote}

This syntax effectively acts as semantic tagging, guiding the LLM to treat certain portions of the JSON string as specialized entities, facilitating a more effective and nuanced parsing of the annotations.

Lastly, we reinforce the \textit{persona} by repeating it and informing the LLM about the \textit{intended usage} of the generated data. We also include any \textit{response restrictions}, which tell LLMs the kind of language to use.

\begin{quote}
\textit{Draw on your professional expertise as a \{role\}, image-specific features mentioned in the annotation, general facts known about the \{target\}, and any other relevant knowledge that can be used to teach \{task\}. \{response restrictions\}.}
\end{quote}

\begin{figure*}[ht]
    \centering
    \begin{subfigure}[b]{0.17\linewidth}
        \includegraphics[width=\linewidth]{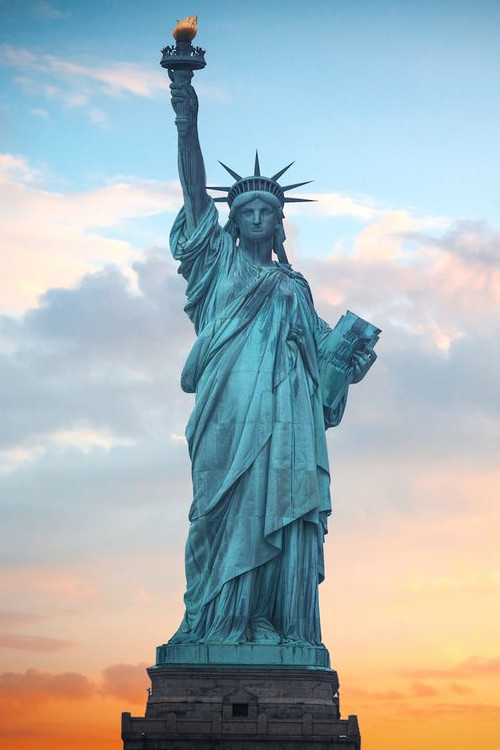}
        \caption{\scriptsize \textbf{The Statue of Liberty} with label "\textbf{holding a torch aloft}.}
        \label{fig:liberty}
    \end{subfigure}
    \hfill
    \begin{subfigure}[b]{0.43\linewidth}
        \begin{minipage}{\linewidth}
        \begin{mdframed}[innerleftmargin=4pt, innerrightmargin=4pt, innertopmargin=4pt, innerbottommargin=4pt]
            \scriptsize
            There is one person in the image who is holding a torch aloft. The person is depicted in \textbf{a classic torch-bearer pose, with their arm raised, suggesting a sense of victory or celebration}.
    
            The person occupying the central portion of the image is standing straight with their right arm held high. 
            
            The right leg is slightly bent at the knee, with the ankle in front of the right hip, \textbf{suggesting a slight forward movement}. The left leg is straight and aligned with the left hip, \textbf{providing stable support to the body}. 
            
            The right arm is positioned in an elevated and bent pose, with the wrist being slightly higher than the shoulder, \textbf{indicative of holding a torch aloft}. The left arm appears relaxed and slightly bent forward at the elbow, hanging naturally by the side. 
            
            The alignment of the pelvis, thorax, and upper neck suggests an upright posture for the torso. The head is also upright, in line with the neck, indicating a forward gaze. \textbf{This overall body posture implies a sense of determination and focus}.
        \end{mdframed}
        \end{minipage}
        \caption{\scriptsize \textbf{Prompt:} Complete prompt with an activity label for the image and keypoint data of the statue.}
        \label{fig:liberty_baseline}
    \end{subfigure}
    \hfill
    \begin{subfigure}[b]{0.36\linewidth}
        \begin{minipage}{\linewidth}
        \begin{mdframed}[innerleftmargin=4pt, innerrightmargin=4pt, innertopmargin=4pt, innerbottommargin=4pt]
            \scriptsize
            There is one person in the image who is standing upright. The person occupies the majority of the center and lower half of the image.
    
            The centrally placed person is standing straight with their limbs showing clear, natural bends and placements. 
            
            Their right leg features the knee slightly bent forward, positioning the right ankle slightly ahead of the right hip. Conversely, their left leg is straight, with the left ankle is directly beneath the hip. 
            
            Their right arm is elevated and bent at the elbow, with the wrist slightly higher than the shoulder. Conversely, the left arm hangs down naturally, with a slight forward bend at the elbow that positions the wrist slightly in front of the shoulder.
            
            The person's torso is upright, as shown by the alignment of the pelvis, thorax, and upper neck. The head is also aligned with the neck, suggesting a forward gaze.
        \end{mdframed}
        \end{minipage}
        \caption{\scriptsize \scriptsize \textbf{Prompt:} Modified prompt containing only keypoint data (i.e., no activity label).}
        \label{fig:liberty_no_activity}
    \end{subfigure}
    \caption{\textbf{Impact of activity labels.} We manually labeled the keypoints on the statue and defined an activity name (a). Using the GPT-4 model, we compare the pose descriptions generated from activity labels and keypoint data (b) with those generated only from keypoint data (c). The additional contextual information included in the LLM response when using the activity label is \textbf{bolded}. These details are absent when the activity label is omitted.}
\end{figure*}

\begin{figure*}[ht]
    \centering
    \begin{subfigure}[b]{0.2\linewidth}
        \includegraphics[width=\linewidth]{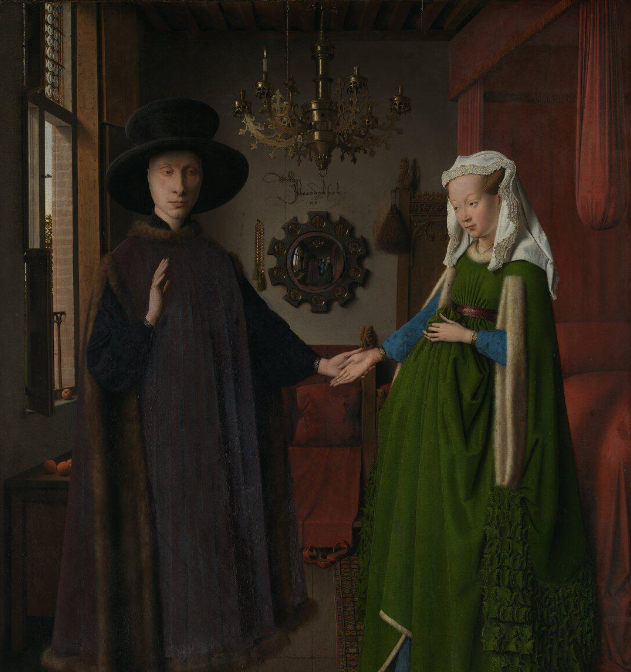}
        \caption{\scriptsize \textbf{The Arnolfini Portrait} shows a married couple. Using the activity label "\textbf{holding hands, indoors}," we demonstrate the effect of persona specification in the LLM prompt when asked to describe this image from keypoints.}
        \label{fig:alnolfini}
    \end{subfigure}
    \hfill
    \begin{subfigure}[b]{0.41\linewidth}
        \begin{minipage}{\linewidth}
        \begin{mdframed}[innerleftmargin=4pt, innerrightmargin=4pt, innertopmargin=4pt, innerbottommargin=4pt]
            \scriptsize
            {\sethlcolor{violet!20}\hl{There are two people in the image who are indoors, holding hands. Both individuals appear to be \textbf{standing upright, facing each other. Their arms are extended towards each other}, \textit{indicating the action of holding hands}. \textbf{Their legs and torsos seem to be in a neutral position}, \textit{suggesting a relaxed state}.}}

            {\sethlcolor{black!20}\hl{The person on the left side of the image is \textbf{standing upright with their body slightly tilted towards the other person. Their \textit{right arm} is extended towards the person on the right, with a slight bend at the elbow}, \textit{indicating they are holding hands}. \textbf{The \textit{left arm} is bent at the elbow with the wrist positioned near the torso. The legs are straight with a slight bend at the knees}, \textit{suggesting a relaxed stance}.}}
            
            {\sethlcolor{green!20}\hl{The person on the right side of the image is \textbf{also standing upright, with a similar body tilt towards the left. Their \textit{left arm} is extended towards the person on the left}, \textit{mirroring the hand-holding action}. \textbf{The \textit{right arm} is bent at the elbow with the wrist positioned near the torso. The legs are straight with a slight bend at the knees}, \textit{mirroring the relaxed stance of the other person}.}}  {\sethlcolor{violet!20}\hl{\textbf{Their heads are slightly inclined towards each other}, \textit{suggesting a sense of intimacy or engagement in conversation. The overall body language suggests a comfortable and relaxed interaction between the two individuals}.}}
        \end{mdframed}
        \end{minipage}
        \caption{\scriptsize \textbf{Prompt:} You are an \underline{expert human activity and pose} \underline{analyzer} with deep understanding of MPII Human Pose dataset, which has 16 keypoints in order \ldots. Given a set of 2D \ldots}
        \label{fig:alnolfini_baseline}
    \end{subfigure}
    \hfill
    \begin{subfigure}[b]{0.38\linewidth}
        \begin{minipage}{\linewidth}
        \begin{mdframed}[innerleftmargin=4pt, innerrightmargin=4pt, innertopmargin=4pt, innerbottommargin=4pt]
            \scriptsize
            {\sethlcolor{violet!20}\hl{There are two people in the image who are holding hands indoors. The individuals are \textbf{standing close to each other, and their hands are connected}, \textit{indicating a shared activity}.}}

            {\sethlcolor{black!20}\hl{The leftmost person is \textbf{standing upright with their body slightly tilted towards the right}. The \textbf{\textit{right leg is straight, with the ankle, knee, and hip aligned vertically}}. The \textbf{\textit{left leg is also straight, but it is slightly angled towards the right}}. The \textbf{torso is upright and slightly turned to the right}. The \textbf{head is straight, looking forward}. The \textbf{right arm is bent at the elbow}, and the \textbf{\textit{hand is extended towards the other person}}. The \textbf{left arm is slightly bent at the elbow}, and the \textbf{hand is also extended towards the other person}, \textit{indicating they are holding hands}.}}
            
            {\sethlcolor{green!20}\hl{The rightmost person is \textbf{also standing upright with their body slightly tilted towards the left}. The \textbf{\textit{right leg is straight, with the ankle, knee, and hip aligned vertically}}. The \textbf{\textit{left leg is also straight, but it is slightly angled towards the left}}. The \textbf{torso is upright and slightly turned to the left}. The \textbf{\textit{head is straight, looking forward}}. The \textbf{right arm is slightly bent at the elbow}, and the \textbf{hand is extended towards the other person}. The \textbf{left arm is bent at the elbow}, and the \textbf{\textit{hand is also extended towards the other person}}, \textit{indicating they are holding hands}.}}
        \end{mdframed}
        \end{minipage}
        \caption{\scriptsize \textbf{Prompt:} Given a set of 2D keypoints from the MPII Human Pose dataset, which has 16 keypoints in order \ldots}
        \label{fig:alnolfini_no_role}
    \end{subfigure}
    \caption{\textbf{Impact of personas.} We manually defined an activity name and labeled keypoints for two people in a sample image (a) and compared GPT-4 output using our prompt, which \underline{specifies a persona in the first sentence} (b) with a modified prompt omitting the LLM role definition (c). For easier comparison, we segment the LLM output into three parts: the first {\sethlcolor{violet!20}\hl{talking about the overall image}}, the second {\sethlcolor{black!20}\hl{talking about the person on the left}}, and the third {\sethlcolor{green!20}\hl{talking about the person on the right}}. The \textbf{text segments describing body pose are bold}, whereas the \textit{text segments drawing insights from the pose are italic}. The \textbf{\textit{incorrect or superfluous pose descriptions are bold-italic}}. When we ask the LLM to act as an expert pose analyzer (b), it makes fewer mistakes, uses more engaging language, and provides higher-quality insights about the pose. Compared to this, when directly asked to describe pose without specifying a role (c), the LLM focuses on insignificant details (i.e., legs, which are not important to the activity), writes monotonic sentences, makes more mistakes, and does not provide useful insights about the interaction.}
    \label{fig:ablation_married_couple}
\end{figure*}

\subsection{Qualitative Ablations for the Prompt}
\label{sec:supp_prompt_ablations}

This section systematically evaluates the impact of different components of our prompt. As described in the paper, our prompt has several essential elements. We look at the response of the GPT-4 model with different prompt variations using the reference images in \ref{fig:liberty} and \ref{fig:alnolfini}. We manually labeled the 16 MPII keypoints for each individual in the reference images and also defined activity labels for each image, which can be found in the respective caption.

\noindent \textbf{Impact of activity labels.} In the paper, we said that supplementing the raw keypoint coordinates with activity labels enhances the context-richness of the generated response. In some cases, it also leads to better parsing of the keypoints. This can be observed by comparing the default description in~\ref{fig:liberty_baseline} with the description in~\ref{fig:liberty_no_activity}. The second description was generated using our complete prompt, with only the activity label missing from the image-specific attributes we provide as key-value pairs. As seen in~\ref{fig:liberty_no_activity}, the model response lacks the additional context provided by activity labels, which are \textbf{marked bold} in ~\ref{fig:liberty_baseline}. This context can enable a better understanding of both the activity and body posture.

Also, it is interesting to note that the activity label can help the LLM infer what objects the person might be interacting with. For example, consider the second image in the third-row of Paper Fig. 4. It shows a person painting a wall with their left arm raised high, similar to the Statue of Liberty. The activity label for this image is ``painting a wall," which produces a widely different (yet accurate) description: ``Their left arm is extended overhead, with their hand holding a tool or brush near the top of the wall."

\noindent \textbf{Impact of personas.} Another principal component of our prompt (\ref{sec:llm_prompt_mpii}) is the persona specification using a \underline{role label}. In~\ref{fig:ablation_married_couple}, we compare the LLM response with and without role labels specifying the persona. When we ask the LLM to act as an expert pose analyzer, it makes fewer mistakes, uses more engaging language, and provides higher-quality insights about the pose. Compared to this, when directly asked to describe pose without specifying a role, the LLM focuses on insignificant details (i.e., legs, which are not important to the activity), writes monotonic sentences, makes more mistakes, and does not provide useful insights about the interaction.

\section{Classification Tasks and Model Details}
\label{sec:tasks_and_datasets}

Given a query image and a set of possible class labels, a task-specific sentence template is populated by the class labels and corresponding text embeddings are computed. The query image is passed through the visual encoder to get the image embeddings. Then, we compute a similarity score between the image embedding and each text embeddings and select the class label corresponding to the most similar text embedding as the predicted class.

We populated a task-specific template sentence with all candidate classes to generate candidate sentences. These sentences were then passed through the pretrained text encoder to obtain text embeddings. Similarly, we passed the query image through the pretrained image encoder to get its image embedding. We used a distance metric to select the text embedding most similar to the image embedding, and we chose the corresponding class label as the predicted class. Below, we provide sentence templates for each task:

\noindent \textbf{Action recognition} aims to predict the action category from images of people performing actions. We use the sentence template: ``\textit{a photo of a person [activity verb-ing]}''.

\noindent \textbf{Age classification} classifies people into a discrete age category using the sentence template: ``\textit{a photo of a [age group] person}''. 

\noindent \textbf{Emotion recognition} predicts the emotion category from images of cropped faces or people. For facial emotion, the sentence template ``\textit{a photo of a/an [emotion adjective] looking face}'' is used. For body images, the sentence template ``\textit{a photo of a person who is feeling [emotion noun]}'' is used.

\subsection{Model Hyperparameters}
\label{sec:hyperparameters}

We provide detailed hyperparameters used to train our models in~\ref{tab:hyperparameters}.

\begin{table}[ht]
    \centering
    \scriptsize
    \caption{FocusCLIP hyperparameters}
    \begin{tabular}{lll}
    \toprule
    & \textbf{Attribute} & \textbf{Value} \\
    \cmidrule(l){1-2} \cmidrule(l){3-3}
    \multirow{7}{*}{\rotatebox{90}{\makecell[t]{\textbf{Training} \\ \textbf{Hyperparams}}}}
    & Batch size & 32 \\
    & Training epochs & 64 \\
    & Optimizer & SGD \\
    & Learning rate &  $1 \times 10^{-3}$ \\
    & Momentum & 0.9 \\
    & Embedding dimension & 512 \\
    & Temperature & 0.5 \\
    \cmidrule(l){1-2} \cmidrule(l){3-3}
    \multirow{6}{*}{\rotatebox{90}{\makecell[t]{\textbf{Vision} \\ \textbf{Transformer}}}}
    & Variant & ViT-Base \\
    & Patch size & 16 \\
    & Input resolution & 224 \\
    & Num. layers & 12 \\
    & Num. heads & 12 \\
    & Hidden dimension & 768 \\
    \cmidrule(l){1-2} \cmidrule(l){3-3}
    \multirow{6}{*}{\rotatebox{90}{\makecell[t]{\textbf{Text} \\ \textbf{Transformer}}}}
    & Variant & BERT-Base-Uncased \\
    & Vocabulary & 30522 \\
    & Num. layers & 12 \\
    & Num. heads & 12 \\
    & Hidden dimension & 768 \\
    & Sequence length & 512 \\
    \bottomrule
    \end{tabular}
    \label{tab:hyperparameters}
\end{table}

\section{Dataset Access}

\textbf{MPII Pose Descriptions Dataset} is accessible through the HuggingFace Datasets API using the corresponding configuration names. We provide up to four pose descriptions for each annotated image, each written by a different LLM.  We used three state-of-the-art LLMs, including OpenAI's GPT-3.5-Turbo and GPT-4, along with the open-source LLaMA-2 model from Meta. For OpenAI models, we used the chat completion endpoint of OpenAI's API. Specifically, we used the \texttt{gpt-3.5-turbo-0613} and \texttt{gpt-4-0613} models. For LLaMA-2, we used the \texttt{llama-2-70b-chat-hf} model with HuggingFace's Inference API. We also used an older GPT-3.5 release, \texttt{gpt-3.5-turbo-0301} with an unrefined version of the prompt from the paper. This model is denoted as GPT-3.5$^*$. Descriptions for the complete training set of MPII~\cite{andriluka14cvpr} are provided for each LLM. The validation set descriptions for GPT-4 are not available.

\end{document}